\documentclass{article}
\usepackage{etex}
\usepackage{nips10submit_e,times}
\usepackage[square,numbers]{natbib}
\usepackage{amsmath, epsfig}
\usepackage{amsfonts}
\usepackage{subfigure}
\usepackage{graphicx}
\usepackage{amsfonts}
\usepackage{algorithm}
\usepackage{algorithmic}
\usepackage{easybmat}
\usepackage{footmisc}
\usepackage{tikz}
	\usepackage{csvsimple}
	\usepackage{pgfplotstable}

\newcommand{\ignore}[1]{}
\newcommand{\comment}[1]{}

\title{Inferring Team Strengths Using a Discrete Markov Random Field}

\author{
John Zech\hspace{4.0cm}Frank Wood\\
\hspace{0.2cm}Windcrest Discovery Investments\hspace{0.85cm}Department of Engineering Science\\
\hspace{-0.6cm}40 Richards Avenue, 3rd Floor\hspace{2.10cm}University of Oxford\\
\hspace{-0.13cm}Norwalk, CT 06854, USA\hspace{2.35cm}Oxford OX1 3PJ, UK\\
\texttt{\hspace{-0.1cm}jzech@wdinvestments.com\hspace{1.1cm}fwood@robots.ox.ac.uk}
}

\nipsfinalcopy

\begin{document}

\maketitle

\begin{abstract}
We propose an original model for inferring team strengths using a Markov Random Field, which can be used to generate historical estimates of the offensive and defensive strengths of a team over time. This model was designed to be applied to sports such as soccer or hockey, in which contest outcomes take value in a limited discrete space. We perform inference using a combination of Expectation Maximization and Loopy Belief Propagation. The challenges of working with a non-convex optimization problem and a high-dimensional parameter space are discussed. The performance of the model is demonstrated on professional soccer data from the English Premier League. 

\end{abstract}

\section{Introduction}
Which is the best team in the English Premier League this year? If one asks five different soccer enthusiasts, one may receive five different answers. Some will highlight the team's realized Win-Draw-Loss record, while others will cite statistical estimates such as ESPN's Power Rankings \citep{Silver:2009:Online}. Everyone agrees that competitors are of varying skill, but establishing how strong any given team is can be difficult, as that team's underlying strength is never directly observed. Poor recent results can always be explained away by a lack of spirit or a tough lineup; good recent results may simply be a lucky run. 

While team strength is unobserved, we do directly observe some statistics about the game. In soccer, we observe the final score, shots on goal, possession statistics, penalties, etc. We might want to use those to infer team strength, but we face a number of challenges. There are no statistics generated exclusively by a given team's offense; we have only statistics generated by that offense in combination with the defense of some other team. That defense, in turn, always appears in combination with yet other offenses. Our estimate of each team's strength depends on our estimate of every other team's strength, and we will need to estimate all of them simultaneously. Further complicating matters, team strengths can (and do) change over time. We might like to have a principled way of updating our beliefs about a team's strength as a season progressed that incorporated both of these facts. 

We propose a model that assumes team offensive and defensive strengths can change over time and that the final scores we observe are generated probabilistically as a function of underlying team strengths. We will look at some previous ranking models that have been applied to soccer, and we will then provide background on two methods from the field of machine learning (Expectation Maximization and Belief Propagation) that will prove useful for developing our model. We then lay out model notation and methodology, explaining how unobserved model parameters and unobserved strengths can both be simultaneously estimated. Finally, we present our estimates of the underlying offensive and defensive strengths of teams in the English Premier League from 1993 to 2012. Additionally, we examine the predictive performance of our model.

\begin{figure*}[ht]
\label{fig:manchester_city_offense}
\begin{center}\includegraphics[scale=0.38]{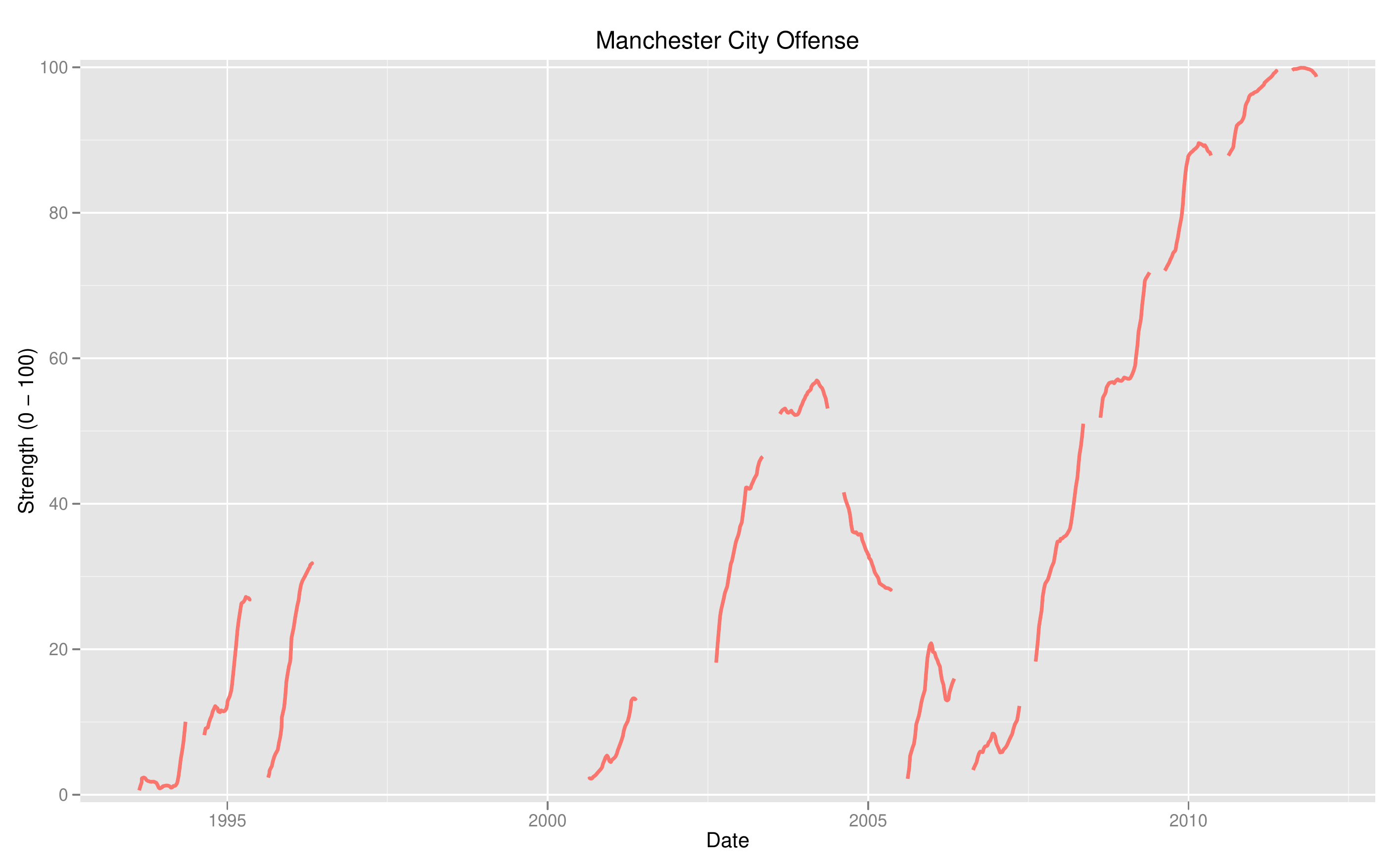}
\caption{Inferred Manchester City offensive strength, 1993-2012. For the purposes of plotting, an expectation was taken over four discrete states (0-3), and this expectation was rescaled to a strength ranging from 0 to 100. Breaks in the line represent time between seasons, which can be relatively short between two successive Premier League seasons or relatively long during periods of relegation (as happened after the 1995-1996 season).}
\end{center}
\end{figure*}

Before we lay out the technical details of the model, we can demonstrate the kind of result it produces. Figure \ref{fig:manchester_city_offense} is a graph of our model's estimate of Manchester City's latent offensive strength from 1993 to 2012. 

The only data available to this model is the number of goals scored by each team in each game they played. As one can see, the model infers a dramatic improvement in Manchester City offensive strength from the beginning of the 2008 - 2009 season onward. This mirrors the September 2008 purchase of the team by Sheikh Mansour bin Zayed Al Nahyan, who spent lavishly on talent and transformed the team from one that went 15-5-18 (wins, draws, and losses, respectively) in 2008-2009 to one that went 28-5-5 in 2011-2012. The model did not have access to the fact of his purchase, but it inferred the evolution in strength regardless based on observed match data.

\subsection{Related Work}
	One of the most widely known models for inferring underlying opponent strength based on the outcome of one-on-one matches is the Elo ranking system. This system was originally created to rank chess players, but has now been applied to rank competitors in other areas such as soccer  \citep{ELOSOCCER} and Scrabble \citep{SCRAB}. In this system, it is assumed that the skill of each contestant is normally distributed with the same variance but possibly different means. The result (win, draw, loss) of a one-on-one contest is simplified to a win or loss (with draws being counted as a half-win, half-loss). This outcome is compared to the previously estimated probability that each player was expected to win or lose given his or her incoming rating. If the player outperforms their expectation, their rating is adjusted upward by an amount proportional to the outperformance and a scaling factor called a ``K-factor." Such a system could be applied to our problem at the team level, treating teams as individual contestants in one-on-one contests.

The original Elo system featured a number of simplifications to allow for easy calculation. Many adaptations of the system have been proposed. One of the most relevant to the problem of modeling soccer is Microsoft's TrueSkill\texttrademark system, which has extended the basic Elo methodology to allow individual player strengths to be estimated in a team setting, such as one might find in the context of online multiplayer gaming \citep{DBLP:conf/nips/HerbrichMG06}. This individually-based ranking approach could be extended to a sport like soccer. An important weakness of this model for our purposes is that we see limited rotation among team members in professional soccer games, as rosters change slowly. If two players often play together, we will not be able to establish how much each contributes individually. This is not a problem in the context for which TrueSkill was designed, as online teams are computer-selected from a large pool of players, and a contestant plays with many different teammates.

There are also ranking systems that move beyond purely mathematical approaches and try to incorporate human intuition for the game into a mathematical framework. One of the most widely followed of these types of models in soccer is the Soccer Power Index (SPI) created by Nate Silver for ESPN. This model is based on observed scoring data, but it makes a number of adjustments for expected competitiveness, home-field advantage, and the recentness of the observation. Both team-level and individual-level strengths are estimated, and those strengths are combined into an overall rating that can be used to predict the outcome of a game. More detail on this approach can be found in \citep{Silver:2009:Online}.

\section{Methodological Background}
Some readers may find a review of Expectation Maximization and Belief Propagation useful, as we rely upon these techniques to do inference in our model. Those already familiar with Expectation Maximization and Belief Propagation are welcome to skip ahead to Section 3. 

\subsection{Expectation Maximization}

The method of maximum likelihood (ML) is used to find parameters for statistical models that match these models to observed data. In ML estimation, one assumes a parametric form for the process that generated the observed data and then selects the parameters $ \hat{\theta} = \{ \hat{\theta}_1,...,\hat{\theta}_{k}\}$ for that distribution under which the observed data are most likely to have occurred. Given observed data points $\mathcal{X}=\{X_1, ..., X_n\}$ and parameters $\theta = \{ \theta_1,...,\theta_{k}\}$, the likelihood function is

{\centering
$ L(\theta|\mathcal{X}) = \displaystyle\prod_{i=1}^{n} P(X_i | \theta)$.\\
}
ML estimation finds $\hat{\theta}$ such that

 \begin{center}
$\hat{\theta}=\underset{\theta}{\operatorname{argmax}}[L(\theta|\mathcal{X})]$.
 \end{center}

This is frequently achieved by taking the derivative of the log-likelihood with respect to the parameters $\theta$ and identifying critical points where the derivative is equal to zero.
One may wish to estimate ML parameters $\hat{\theta}$ in a model in which there are latent variables $\mathcal{Z}$ as well as observed variables $\mathcal{X}$ and parameters $\theta$. 
We define $\mathcal{Z} = \{ Z_1,...,Z_k \}$, where $Z_1,...,Z_k$ are variables that take values in different spaces of the latent variables. Given latent variables, the likelihood above can be written

 \begin{center}
$P(\mathcal{X}|\theta) = \displaystyle\sum_{\mathcal{Z}}P(\mathcal{X},\mathcal{Z}|\theta) = \displaystyle\sum_{\mathcal{Z}}[\prod_{i=1}^{n} P(X_i,\mathcal{Z}|\theta)].$
 \end{center}

It will be considerably more difficult to take the derivative of this function with respect to individual parameters, as the logarithm of the log-likelihood will not pass through the summation over $\mathcal{Z}$ to the product over $\mathcal{X}$. To perform ML estimation of $\theta$ in models with latent variables $Z$, we make use of the EM algorithm, first introduced by Dempster, Laird, and Rubin \citep{Dempster77maximumlikelihood}. A good textbook introduction can be found in Bishop \citep{Bishop:2006:PRM:1162264}. Formally, the EM algorithm seeks $\hat{\theta}$ such that 
 \begin{center}
$\hat{\theta} = \underset{\theta}{\operatorname{argmax}}   [Q(\theta,\theta^{old})]  $
 \end{center}

 \begin{center}
 where
\begin{equation}
Q(\theta,\theta^{old})=\displaystyle\sum_{\mathcal{Z}} [P(\mathcal{Z}|\mathcal{X},\theta^{old}) \mathrm{ln}(P(\mathcal{X},\mathcal{Z}|\theta))]. \label{eq:no1}
\end{equation}
\\

 \end{center}

This algorithm is guaranteed to generate an estimate $\hat{\theta}$ for $\theta$ that converges to a local maximum of the likelihood. The EM algorithm may be understood as repeating two steps, M and E. In step 1 (M step), one holds a learned distribution over $Z$ constant and maximizes $Q$ with respect to  $\theta$, storing this as $\theta^{old}$. In step 2 (E step), one uses the estimate of $\theta^{old}$ to compute a distribution over $Z$. These two steps are repeated sequentially until $\theta$ converges to $\hat{\theta}$. For more detail, please refer to Bishop \citep{Bishop:2006:PRM:1162264}.

\subsection{Belief Propagation}
Belief propagation (BP) is a technique that can be used to efficiently calculate marginal distributions in large models with many variables. It was introduced by Pearl \citep{DBLP:conf/aaai/Pearl82}, and a good textbook treatment can be found in Bishop \citep{Bishop:2006:PRM:1162264}. In the context of EM, it can be used to compute $P(\mathcal{Z}|\mathcal{X},\theta^{old})$.

We review BP for a simple case in which there is a joint distribution over $N$ discrete variables $\{X_1,...,X_N\}$, each of which is discrete and can take $K$ possible values.\footnote{BP may also be used when the underlying latent variables are not discrete, although the mathematics become more complicated as integrals replace summations. Weiss et. al. \citep{DBLP:conf/nips/YedidiaFW00} provides an extension of BP to Gaussian latent variables.  } We assume that the joint may be factorized

\begin{center}
$P(X_1,...,X_N) = P(X_1)P(X_2|X_1)P(X_3|X_2)...P(X_i|X_{i-1})P(X_{i+1}|X_i)...P(X_{N}|X_{N-1})$.
\end{center}

Instead of considering the entire distribution, we may want to investigate the distribution of one or several variables, ignoring all others. We can do this by marginalizing over the variables we wish to ignore. For example, if one wanted to find the distribution over a single $X_i$ such that

\begin{center}
$P(X_i)=\displaystyle\sum_{X_1} \displaystyle\sum_{X_2}... \displaystyle\sum_{X_{i-1}} \displaystyle\sum_{X_{i+1}} ... \displaystyle\sum_{X_n} P(X_1,...,X_n)$.
\end{center}

\begin{figure}[h]
  \centering
  \includegraphics[scale=1.1]{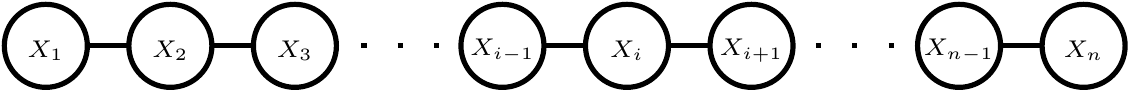}
  \caption{Graphical representation of factorized joint distribution}
  \label{fig:bpgraph}
\end{figure}

This factorization is represented graphically by Figure~\ref{fig:bpgraph}. Each variable is referred to as a ``node'' of the graph, and lines connecting nodes indicate dependencies between variables. 
\\
\\

One may compute this marginal by evaluating all $K^N$ possible states of the joint distribution and summing over them appropriately. The computational cost of this calculation will scale exponentially with $N$, the number of variables in the distribution, and so is of $O(K^N)$. This will become computationally infeasible as $N$ becomes moderately large.

Belief propagation offers a computationally advantageous way to calculate marginal distributions. First, we substitute the factorized joint distribution for the full joint distribution

\begin{center}
$P(X_i)=\displaystyle\sum_{X_1} \displaystyle\sum_{X_2}... \displaystyle\sum_{X_{i-1}} \displaystyle\sum_{X_{i+1}} ... \displaystyle\sum_{X_n}                P(X_1)P(X_2|X_1)P(X_3|X_2)...P(X_i|X_{i-1})P(X_{i+1}|X_i)...P(X_{n}|X_{n-1})$.
\end{center}

Terms may be reordered as long as each term is kept to the right of the summation over it, i.e.,
\begin{align}\label{eq:reorder}
P(X_i)    &=    [\displaystyle\sum_{X_{i-1}} P(X_i|X_{i-1})...[\displaystyle\sum_{X_2} P(X_3|X_2)[ \displaystyle\sum_{X_1} P(X_2|X_1)P(X_1)]]   \notag \\ 
             &\qquad [\displaystyle\sum_{X_{i+1}} P(X_{i+1}|X_i)... [\displaystyle\sum_{X_{n}} P(X_n|X_{n-1})]]. 
\end{align}

This reordering significantly reduces computational requirements. This calculation now requires $(N-1)$ summations, each of which summations occurs over a $K \times K$ joint distribution. The complete marginalization has computational cost $O(NK^2)$, and thus the number of required operations scales linearly with $N$ rather than exponentially with $N$. In models with large $N$, this is a dramatically reduces the computational cost and makes marginalization computationally feasible. 

There is another interpretation of Formula~\ref{eq:reorder}  for the marginal $P(X_i)$. Moving from right to left, one may see the top line as a ``message'' sent from $X_1$ to $X_i$, and may see the bottom line as a ``message'' sent from $X_N$ to $X_i$. One starts at one end of the chain, say $X_1$. One marginalizes over $X_1$ and stores the resulting distribution over $X_2$ as $M_1$. One then moves down the chain to the next variable, $X_2$, piecewise multiplies by the previously calculated $M$ (here, $M_1$), marginalizes over $X_2$, and stores the result as $M_2$. One continues on until one reaches $X_i$, where one stops. This process is repeated from the other end of the chain. In this way, the sending of messages down chains can be used to complete marginalizations. One advantage of seeing belief propagation as message passing is that it highlights the local nature of the calculation. Once one has calculated and stored message $M$ above at each point, it is possible to calculate marginals for any node in the graph by using only the messages immediately feeding into that node. We will take advantage of this fact to efficiently calculate marginal distributions later on.

\begin{center}
\begin{figure*}[h]
  \includegraphics[scale=0.98]{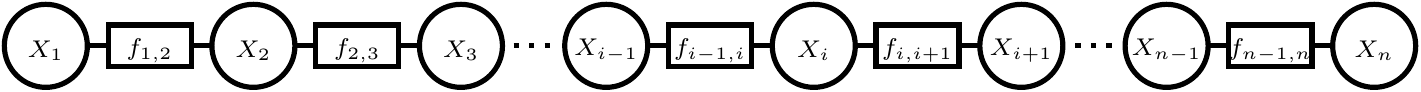}
  \caption{Factor graphical representation\label{fig:bpfactorgraph}}
\end{figure*}
  \end{center}

The above graph may also be interpreted as a product of ``factors'' which encode the joint distribution between neighboring variables. A factor representation of the above chain is given in Figure \ref{fig:bpfactorgraph}. In this figure, one can see that each pair of neighboring variables is governed by the same factor. For example, $X_2$ and $X_3$ are connected by $f_{2,3}$. This signals that $X_2$ and $X_3$ are governed by the factor $f_{2,3}$, which encodes the relationship $P(X_3|X_2)$. Factor graphs are a convenient representation to employ when doing message passing.

The above discussion assumes that the joint distribution $P(X_1,...,X_N)$ factors as described. The same argument may be extended to ``tree-shaped'' graphs, in which there is only a single path from one node to any other. Certain graphs, however, will not factor in this way. In this case, belief propagation may still be used to generate approximate marginals, but it is no longer guaranteed converge to a correct result. We make use of loopy BP for the E step of parameter learning in our model.\footnote{For more information on loopy belief propagation, please see Frey and MacKay \citep{DBLP:conf/nips/FreyM97} or section 8.4.7 of Bishop  \citep{Bishop:2006:PRM:1162264}.}

\section{Model (Notation)}

We begin with a number of definitions. 
 Let $\mathcal{D}= \{1,...,D\}$ be an ordered set. These will be a sequence of dates. 
Let $d \in \mathcal{D}$ be a variable taking values in the ordered set $\mathcal{D}$. Let $\mathcal{T}=\{1,...,T\}$ be a set; these will refer to teams. Let $t \in \mathcal{T}, a \in \mathcal{T}, h \in \mathcal{T}$ be variables taking values in this set. Let  $ I_{h,a,d}$, $h \in \mathcal{T}, a \in \mathcal{T}, d \in \mathcal{D}$, be an indicator function defined such that

\begin{center}
$
I_{h,a,d} =
\begin{cases} 
1 &\text{if home team $h$ and away team $a$ played on date $d$}  \\
0 &\text{otherwise} .
\end{cases}
$
\end{center} 

\noindent Let $\mathcal{G}=\{1,...,G_{max}\}$ be an ordered set of indices referring to possible goal totals.\footnote{To reduce model complexity, goals scored are capped at 4. Teams rarely score more than this number of goals (less than 2\% of the time in our English Premier League data).}

Let $M_{h,a,d,g}$ be an indicator function such that\\

\begin{center}
$
M_{h,a,d,g}
\begin{cases} 
1 &\text{if home team $h\in\mathcal{T}$ faced away team $a\in\mathcal{T}$ on date $d\in\mathcal{D}$ and scored $g\in\mathcal{G}$ goals} \\
0 &\text{otherwise} .
\end{cases}
$
\end{center}

\noindent Let $\mathbf{m}_{h,a,d} = \{M_{h,a,d,0}, ..., M_{h,a,d,G_{max}}\}$ be the indicator vector for the goals scored by home team $h \in \mathcal{T}$ facing away team $a \in \mathcal{T}$ at date $d \in \mathcal{D}$. For example, if home team 1 ($t=1$) played away team 2 ($t=2$) at the third date of the sample ($d=3$) and scored four goals ($g=4$) , then $\mathbf{m}_{1,2,3} =\{0,0,0,0,1\}$. Let $\mathcal{M} = \{\mathbf{m}_{1,2,0}, ..., \mathbf{m}_{T,T,D}\}$.

Likewise, let $N_{a,h,d,g}$ be an indicator function such that\\

\begin{center}
$
N_{a,h,d,g}
\begin{cases} 
1 &\text{if away team $a\in\mathcal{T}$ faced home team $h\in\mathcal{T}$ on date $d\in\mathcal{D}$ and scored $g\in\mathcal{G}$ goals} \\
0 &\text{otherwise}.
\end{cases}
$
\end{center}

\noindent Let $\mathbf{n}_{a,h,d} = \{N_{a,h,d,0}, ..., N_{a,h,d,G_{max}}\}$ be the indicator vector for the goals scored by away team $a \in \mathcal{T}$ facing home team $h \in \mathcal{T}$ at date $d \in \mathcal{D}$. Let $\mathcal{N} = \{\mathbf{n}_{1,2,0}, ..., \mathbf{n}_{T,T,D}\}$.

\begin{figure*}[h]
  \centering
  \includegraphics[scale=0.83]{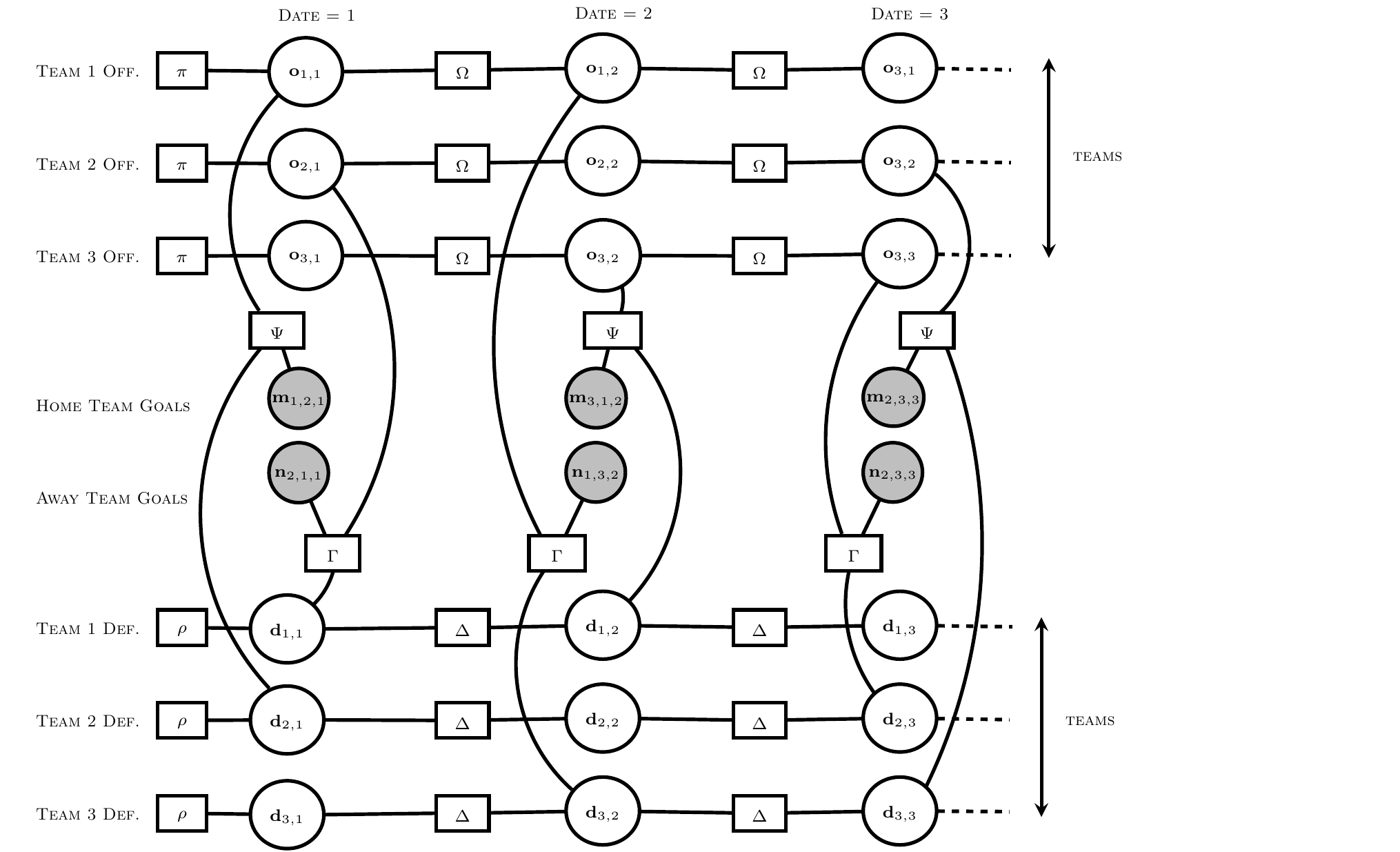} 
  \caption{Factor graph representation of model\label{fig:modelfactorgraph}}
\end{figure*}

With these indicator functions defined, we can develop our model for team strength. As we proceed, it may be helpful to refer to Figure~\ref{fig:modelfactorgraph}, a graphical example of the model. All observed variables are represented by shaded nodes, and unobserved variables are represented by unshaded nodes. This example shows a round robin tournament in which team 1 (home) plays team 2 (away) at time 1, team 1 (home) plays team 3 (away) at time 2, and team 2 (home) plays team 3 (away) at time 3. In each of the three time periods, only one contest occurs.\footnote{The model setup created by the sample EPL data is much larger: there are 45 teams, 678 weeks of data, and an average of approximately 11 contests in any given period.} In every period, each team is assumed to have some latent offensive strength $o \in \mathcal{S}$ and some latent defensive strength $d \in \mathcal{S}$. We cannot directly observe these strengths, which are represented as unshaded circles. We assume that the offensive strength a team occupies can change over time, and that the probability of a team's offensive strength moving from one level to another is governed by an offensive transition matrix. The same is assumed to be true of defensive strength. While underlying strengths are unobserved, we do observe home and away goal totals, which are represented graphically by shaded circles. We assume that a home goal total is determined probabilistically by the home team's offensive strength and the away team's defensive strength. Likewise, we assume that an away goal total is determined probabilistically by the away team's offensive strength and the home team's defensive strength. The aim of the model is to use the observed goals to infer the unobserved underlying offensive and defensive strengths at each point in time.

Let 

\begin{center}
$O_{t,d,s} =
\begin{cases} 
1 &\text{if the offense of team } t \in \mathcal{T} \text{ at date } d\in \mathcal{D} \text{ was at strength } s \in \mathcal{S}  \\
0 &\text{otherwise} .
\end{cases}
$
\end{center}

\noindent Let $\mathbf{o}_{t,d} = \{O_{t,d,1}, ..., O_{t,d,S}\}$ be the indicator vector for the offensive strength of team $t \in \mathcal{T}$ at date $d \in \mathcal{D}$. For example, if the offense of team 1 ($t=1$) at the beginning of the sample ($d=1$) was at offensive strength 3 ($s=3$), then $\mathbf{o}_{1,1} =\{0,0,1,0\}$. Let $O = \{\mathbf{o}_{1,1}, ..., \mathbf{o}_{T,D}\}$.

Likewise, let 

\begin{center}
$D_{t,d,s} =
\begin{cases} 
1 &\text{if the defense of team } t \in \mathcal{T} \text{ at date } d\in \mathcal{D} \text{ was at strength } s \in \mathcal{S}  \\
0 &\text{otherwise} .
\end{cases}
$
\end{center}

\noindent Let $\mathbf{d}_{t,d} = \{D_{t,d,1}, ..., D_{t,d,S}\}$ be the indicator vector for the defensive strength of team $t \in \mathcal{T}$ at date $d \in \mathcal{D}$. Let $D = \{\mathbf{d}_{1,1}, ..., \mathbf{d}_{T,D}\}$.

Let $\Omega$ be an $S \times S$ transition matrix, where $\Omega_{i,j}$ equals the probability of a team moving from offensive strength $i$ to offensive strength $j$ in the subsequent week. 
Let $\Delta$ be an $S \times S$ transition matrix, where $\Delta_{i,j}$ equals the probability of a team moving from defensive strength $i$ to defensive strength $j$ in the subsequent week. 
Let $\pi$ be a vector of length $S$ which gives the initial distribution over offensive strengths.
Let $\rho$ be a vector of length $S$ which gives the initial distribution over defensive strengths. 

Let $\Psi$ be an $S \times S \times G$ emission conditional probability table (CPT) consisting of entries $\Psi_{i,j,g}$, $i \in \mathcal{S}$, $j \in \mathcal{S}$, $g \in \mathcal{G}$. Each entry  $\Psi_{i,j,g}$ represents the probability of the home team at offensive strength $i$ scoring $g$ goals when facing an away defense at strength $j$.
Let $\Gamma$ be an $S \times S \times G$ emission CPT consisting of entries $\Gamma_{i,j,g}$, $i \in \mathcal{S}$, $j \in \mathcal{S}$, $g \in \mathcal{G}$. Each entry $\Gamma_{i,j,g}$ represents the probability of the away team at offensive strength $i$ scoring $g$ goals when facing an home defense at strength $j$. Each of these emission CPTs is of dimension $S \times S \times G$ as it encodes a relationship between offensive strengths (of which $S$ are possible), defensive strengths (of which $S$ are possible), and goal states (of which $G$ are possible).

This model contains a large number of parameters relative to the available data. As a result, ML estimation will tend to fit models that generalize poorly. To address this issue, we will establish a prior distribution for certain parameters and find the most likely parameter estimates considering both the prior and the data. This is called maximum a posteriori (MAP) estimation. We use a Dirichlet prior for both our transition parameters $\Omega$ and $\Delta$ and our goal emission parameters $\Psi$ and $\Gamma$. Our Dirichlet transition parameter is $A$, which consists of $k$ vectors $\mathbf{\alpha_k}$, each of which consists of positive reals $\alpha_{k,1},...,\alpha_{k,j}$, where $k=j=S$. We believe teams are more likely to remain in their current state than to change states. Accordingly, we set $\alpha_{j,k}$ higher if $j=k$.\footnote{It should also be noted that, in our implementation, we make use of two different types of transitions: one within-season and one between-seasons. Noting these separately adds additional complication to the notation and is omitted from this presentation; it is a straightforward extension of the model presented here. For the within-season Dirichlet prior, each $\alpha_{j,k} =  C  ((1.25 S - 0.25)(8^D))$, where $C$ is a number chosen to affect the influence of the regularization, $S$ is the max number of states in the model, and $D=|k-j|$ if $|k-j|<=1$ and $0$ otherwise. For the between-season Dirichlet prior, each $\alpha{j_k} = C/((F)(2^{|k-j|}))$, where  $F = \displaystyle \sum_{i = 1}^{S-1}2 (S-i) / (2^i)$.} Our Dirichlet home goal emission parameter is $\mathbf{\beta}$, a vector of positive reals $\beta_{0},...,\beta_{G}$. Our Dirichlet away goal emission parameter is $ \phi$, a vector of positive reals $\phi_{0},...,\phi_{G}$. We set $\beta_g$ using the distribution over home goals seen historically across professional soccer leagues, and we set $\phi_g$ similarly for away teams. For example, if an away team typically scored zero goals half of the time, we would set $\phi_0=C/2$, where $C$ is a number chosen to affect the influence of the regularization.

In this model,
the set of all observed variables is $\mathcal{X} = \{\mathcal{M},\mathcal{N}\}$, the set of all latent variables is $\mathcal{Z} = \{O,D\}$, the set of all hyperparameters is $ \Lambda = \{ A, \beta, \phi\}$, and the set of all parameters is $\mathbf{\theta} = \{\pi,\rho,\Omega,\Delta,\Psi,\Gamma\}$

\subsection{Methodology}
Our goal is to find parameters $\theta$ that reflect how the offensive and defensive strengths of teams evolve over time, and that capture how likely a team at a given offensive strength is to score a certain number of goals when facing a team at a given defensive strength. We would like to use those learned parameters $\theta$ to discover the strength of teams' offenses and defenses in the past, and, potentially, to predict the outcome of matches between teams in the future. We will infer $\theta$ by using a two step process. First, we will use loopy belief propagation to find the distribution over the latent $\mathcal{Z}$ given the current parameter settings $\theta$ (E step). Then, we will use the M step to update our estimate of $\theta$.

The joint distribution $P(\mathcal{Z},\mathcal{X},\mathcal{\theta} | \Lambda)$ can be factorized as
 \begin{center}
$P(\mathcal{X},\mathcal{Z}, \mathcal{\theta}|\Lambda) = P(\mathcal{Z}|\theta) P(\mathcal{X}|\mathcal{Z},\theta) P(\mathcal{\theta} | \Lambda) $\\

\end{center}
where\\
\begin{align*}
 P(\mathcal{Z}|\theta) &=        \displaystyle\prod_{t \in \mathcal{T}} \{\prod_{i \in \mathcal{S}}(\pi_{i}^{O_{t,1,i}})  (\rho_{i}^{D_{t,1,i}})  (\displaystyle\prod_{d \in \mathcal{D} \setminus 1}\displaystyle\prod_{k \in \mathcal{S}}[(\Omega_{i,k}^{(O_{t,d-1,i}) (O_{t,d,k})})          (\Delta_{i,k}^{(D_{t,d-1,i}) (D_{t,d,k})})         ]\},       
\\  P(\mathcal{X}|\mathcal{Z},\theta)&=\displaystyle\prod_{d \in \mathcal{D}} \displaystyle\prod_{h \in \mathcal{T}} \displaystyle\prod_{a \in \mathcal{T} }\displaystyle\prod_{i \in \mathcal{S}}\displaystyle\prod_{j \in \mathcal{S}} \displaystyle\prod_{g \in \mathcal{G}}\{   \Psi_{i,j,g}^{(M_{h,a,d,g}) (O_{h,d,i})(D_{a,d,j})}  \Gamma_{i,j,g}^{(N_{a,h,d,g}) (O_{a,d,i}) (D_{h,d,j})}                           \}, 
\\ P(\mathcal{\theta} | \Lambda)&\propto \displaystyle\prod_{k\in\mathcal{S}} \displaystyle\prod_{j\in\mathcal{S}}\Omega_{k,j}^{\alpha_{j,k}-1}\Delta_{k,j}^{\alpha_{j,k}-1}     \displaystyle\prod_{k\in\mathcal{S}} \displaystyle\prod_{j\in\mathcal{S}} \displaystyle\prod_{g\in\mathcal{G}}\Psi_{k,j,g}^{\beta_g-1}\Gamma_{k,j,g}^{\phi_g-1}.
\end{align*}

It may be helpful to note that many of the products in the exponents evaluate to zero. We use unnecessary parentheses in many places where we believe it helps readability.

The above can also be expressed as a product of the factors labeled in Figure~\ref{fig:modelfactorgraph}. As before, each circle on the graph represents a variable, and each box on the graph represents a dependency between variables. Each offensive chain begins with $\pi$, a factor that encodes the initial distribution over offensive states, and each defensive chain begins with $\rho$, a factor that encodes the initial distribution over defensive states. Each offensive node is connected to its neighbor by $\Omega$, the offensive transition matrix, and each defensive node is connected to its neighbor by $\Delta$, the defensive transition matrix. $\Psi$ encodes the relationship between offensive strength, defensive strength, and number of home goals scored. $\Gamma$ encodes the relationship between offensive strength, defensive strength, and number of away goals scored. Each factor will return a value that reflects the current setting of its neighbors. For example, if offensive strength $\mathbf{o}_{1,1}$ were set to (1,0,0,0), defensive strength $\mathbf{d}_{2,1}$ were set to (0,1,0,0), and $\mathbf{m}_{1,2,1}$ were set to (0,0,0,1,0) then the factor $\Psi$ attached to those three nodes would return the value $\Psi_{1,2,3}$, which in our model is the conditional probability of seeing 3 home goals given a home offense at strength 1 facing an away defensive at strength 2. By taking the product of all the factors on the graph, we can thus calculate the likelihood of the given state of all variable nodes.

We wish to find the maximum a posteriori parameter estimates $\theta$ given observed values of $\mathcal{X}$ and unobserved latent variables $\mathcal{Z}$, which we will do using the EM algorithm.
We define the following:

$$\vbox{\openup1\jot\halign{#&&\ \hfil$\displaystyle#$\hfil\cr
  &         \gamma_o(O_{t,d,k})    	  		 &=&       \mathbb{E} [O_{t,d,k}]      			&=&    		P(O_{t,d,k}=1|\mathcal{X},\theta,\Lambda)	      \cr
  &         \gamma_d(D_{t,d,k})         		 &=&       \mathbb{E} [D_{t,d,k}]      			&=&     	P(D_{t,d,k}=1|\mathcal{X},\theta,\Lambda)      \cr
  &         \xi_o(O_{t,d-1,j},O_{t,d,k})    	   	 &=&       \mathbb{E} [(O_{t,d-1,j}) (O_{t,d,k})]      	&=&    	P(O_{t,d-1,k}=1,O_{t,d,j}=1|\mathcal{X},\theta,\Lambda)      \cr
  &         \xi_d(D_{t,d-1,j},D_{t,d,k})          	 &=&       \mathbb{E} [(D_{t,d-1,j}) (D_{t,d,k})]       	&=&     	P(D_{t,d-1,k}=1,D_{t,d,j}=1|\mathcal{X},\theta,\Lambda)      \cr
  &         \zeta(O_{t_1,d,k},D_{t_2,d,j})	 &=&       \mathbb{E} [(O_{t_1,d,k}) (D_{t_2,d,j})]   &=&    	P(O_{t_1,d,k}=1,D_{t_2,d,j}=1| \mathcal{X},\theta,\Lambda)      \cr 
}}$$

$\,$\\

The above give distributions over each individual offensive node, each individual defensive node, each subsequent pair of offensive nodes, each subsequent pair of defensive nodes, and each opposing pair of offensive and defensive nodes that interact in creating a goal total. 

The above expectations require the distribution over $\mathcal{Z}$ as calculated using BP in the E step of the EM algorithm. A example of how this distribution can be calculated from messages is given as follows. Suppose that four strength states are possible, and that the offensive strength of team $t$ at time $d$ has an incoming message from its past offensive strength at time $d-1$ of (0.1, 0.2, 0.3, 0.4), an incoming message from its future offensive strength at time $d+1$ of (0.14, 0.16, 0.32, 0.38), and an incoming message from a home goal node at time $d$ of (0.45, 0.25, 0.15, 0.05). This represents an offense that is relatively strong at times $d-1$ and $d+1$, with more probability mass on higher states, but that has performed poorly in the game at time $d$. The distribution over the team's offensive strength state at this time $d$ can be obtained simply by multiplying these three vectors in element-wise fashion and rescaling the result so that the distribution sums to one. Here, the latent offensive strength would be calculated to be ((0.1)(0.14)(0.45), (0.2)(0.16)(0.25), (0.3)(0.32)(0.15), (0.4)(0.38)(0.05)) = (0.0063, 0.008, 0.0144, 0.0076). We rescale this vector to sum to one and find a latent offensive strength distribution at time $t$ of (0.174, 0.220, 0.397, 0.209). In the earlier described notation, we would write that $\gamma_o(O_{t,d,0})$=0.174, $\gamma_o(O_{t,d,1})$=0.220, $\gamma_o(O_{t,d,2})$=0.397, and $\gamma_o(O_{t,d,3})$=0.209.

The same procedure can be used to find the distribution over goals. In the interest of clarity, we will suppose that only two strength states are possible (0 and 1) and that only two goal states are possible (0 and 1). Suppose that a mediocre home offense sending message (0.7, 0.3) encounters a strong away defense sending message (0.1, 0.9). These two messages encounter each other at a goal factor $\Psi$ with factor values as given in the conditional probability table in Figure~\ref{fig:psi_example}.

\begin{figure*}[ht]
\begin{center}
\begin{filecontents*}{psi_mp_example.csv}
g:,0,1
$\Psi_{0,0,g}$,0.55,0.45
$\Psi_{0,1,g}$,0.75,0.25
$\Psi_{1,0,g}$,0.2,0.8
$\Psi_{1,1,g}$,0.45,0.55
\end{filecontents*}
\csvautotabular{psi_mp_example.csv}
\caption{Example parameters for $\Psi$ used to illustrate details of message passing.  \label{fig:psi_example}}
\end{center}
\end{figure*}

The first line of the conditional probability table for $\Psi$ in Figure~\ref{fig:psi_example} can be understood as encoding the belief that when a home offense in strength 0 encounters an away defense in strength 0, that home team will score zero goals 55 percent of the time, and one goal 45 percent of the time. In this example, one can see that stronger offenses tend to score a goal more often, whereas stronger defenses tend to prevent goals from being scored. 

We can combine this $\Psi$ factor with the incoming messages to generate a distribution over goals. Practically, this calculation involves multiplying every entry $\Psi_{x,y}$ with the corresponding components of incoming messages from an offensive state $x$ and a defensive state $y$ and then marginalizing out to get a distribution over the goals. This calculation is illustrated in Figure~\ref{fig:cpt_calc}

 \begin{figure*}[ht]
\begin{center}
\begin{filecontents*}{cpt_calc.csv}
$(h) (a) (g)$:,$(h) (a) (0)$,$(h) (a) (1)$
$(h) (a) (\Psi_{0,0,g})$,(0.7)(0.1)(0.55),(0.7)(0.1)(0.45)
$(h) (a) (\Psi_{0,1,g})$,(0.7)(0.9)(0.75),(0.7)(0.9)(0.25)
$(h) (a) (\Psi_{1,0,g})$,(0.3)(0.1)(0.2),(0.3)(0.1)(0.8)
$(h) (a) (\Psi_{1,1,g})$,(0.3)(0.9)(0.45),(0.3)(0.9)(0.55)
,,
Result:,,
$(h) (a) (\Psi_{0,0,g})$,0.0385,0.0315
$(h) (a) (\Psi_{0,1,g})$,0.4725,0.1575
$(h) (a) (\Psi_{1,0,g})$,0.0060,0.0240
$(h) (a) (\Psi_{1,1,g})$,0.1215,0.1485
,,
Sum:,0.6385,0.3615


\end{filecontents*}
\csvautotabular{cpt_calc.csv}
\caption{Example calculation of distribution over goals given $h$, $a$, and $\Psi$ \label{fig:cpt_calc}}
\end{center}
\end{figure*}

This yields a distribution over goals: 63.85 percent of the time we expect to see zero goals scored by the home team, 36.15 percent of the time we expect to see one goal scored by the home team. In this example, the vector over goals we found by summing the two columns itself summed to one, as is required for a valid probability distribution. Usually this vector will not sum to one, which can be remedied in the same way as in the earlier message passing example by rescaling the vector to sum to one.

Given these expectations we can derive parameter updates. We demonstrate by deriving the update for $\Omega_{x,y}$, the probability of moving from offensive strength $x \in \mathcal{S}$ to offensive strength $y \in \mathcal{S}$. Please note that this derivation is repeated for all parameters, as we are optimizing via coordinate ascent. We want to maximize Eqn.~\eqref{eq:no1} with respect to $\Omega_{x,y}$ subject to the constraint that $\displaystyle\sum_{k \in \mathcal{S}}\Omega_{x,k}=1$, as $ \Omega$ is a CPT and the probability of moving from a given state to all possible states must sum to one. Using a Lagrange multiplier, the function $F$ to be maximized is given
\begin{center}
$F(\Omega_{x,k},\Omega_{x,k}^{old})=\displaystyle\sum_{Z \in \mathcal{Z}} [P(Z|X,\Omega_{x,k}^{old}) \mathrm{ln}(P(X,Z,\Omega_{x,k}|\Lambda))] - \lambda(\displaystyle\sum_{k \in \mathcal{S}}\Omega_{x,k}-1)$
\end{center}
where
\begin{align*}
\mathrm{ln}(P(X,Z,\Omega_{x,y} | \Lambda))        &=          \displaystyle\sum_{t \in \mathcal{T}}    \left(     \sum_{i \in \mathcal{S}}       \left(              (O_{t,1,i})\mathrm{ln}(\pi_{i})+(D_{t,1,i})\mathrm{ln}(\rho_{i}) \right)                                						    \right.														\\
                                   &\qquad   + \left.  \left(    \displaystyle\sum_{d \in \mathcal{D} \setminus 1}\displaystyle\sum_{k \in \mathcal{S}}  \left(    ({(O_{t,d-1,i}) (O_{t,d,k})}\mathrm{ln}(\Omega_{i,k}))    +      ({(D_{t,d-1,i}) (D_{t,d,k})}\mathrm{ln}(\Delta_{i,k}) \right)         \right)      \right)            \\
                                   &\qquad  +\displaystyle\sum_{d \in \mathcal{D}} \displaystyle\sum_{h \in \mathcal{T}} \displaystyle\sum_{a \in \mathcal{T} } \displaystyle\sum_{i \in \mathcal{S}}\displaystyle\sum_{j \in \mathcal{S}} \displaystyle\sum_{g \in \mathcal{G}} \left(  \mathrm{ln}(\Psi_{i,j,g}){ (M_{h,a,d,g}) (O_{h,d,i})(D_{a,d,j})} \right. \\
&\qquad \left. +~\mathrm{ln}(\Gamma_{i,j,g}){(N_{a,h,d,g}) (O_{a,d,i}) (D_{h,d,j})}            \right) 
\\ &\qquad +\displaystyle\sum_{k\in\mathcal{S}} \displaystyle\sum_{j\in\mathcal{S}}((\alpha_{j,k}-1)\mathrm{ln}(\Omega_{k,j})+(\alpha_{j,k}-1) \mathrm{ln}(\Delta_{k,j}))     \\ &\qquad +\displaystyle\sum_{k\in\mathcal{S}} \displaystyle\sum_{j\in\mathcal{S}} \displaystyle\sum_{g\in\mathcal{G}}((\beta_g-1)\mathrm{ln}(\Psi_{k,j,g})+(\phi_g-1)\mathrm{ln}(\Gamma_{k,j,g})) - K
\end{align*}

where K is a normalization constant introduced by the Dirichlet priors. We are maximizing the above $\mathrm{ln}(P(X,Z,\Omega_{x,y}|\Lambda))$ with respect to a single parameter, $\Omega_{x,y}$. To maximize the above with respect to $\Omega_{x,y}$, we will take a derivative with respect to $\Omega_{x,y}$. Since most of the terms in the above are constants with respect to $\Omega_{x,y}$, they will effectively fall into the constant term when we take that derivative, and so we may rewrite the function to be optimized as
\begin{align*}
F(\Omega_{x,y},\Omega_{x,y}^{old}) &= \mathbb{E}[\mathrm{ln}(P(X,Z,\Omega_{x,y}|\Lambda))] - \lambda(\displaystyle\sum_{k \in \mathcal{S}}\Omega_{x,k}-1) \\ &=  \mathbb{E}[\mathrm{ln}(\Omega_{x,y})[(\alpha_{x,y}-1)+(\displaystyle\sum_{t \in \mathcal{T}}\displaystyle\sum_{d \in \mathcal{D} \setminus 1}O_{t,d-1,x}O_{t,d,y})]]  +  C -  \lambda(\displaystyle\sum_{k \in \mathcal{S}}\Omega_{x,k}-1)
\\&= \mathrm{ln}(\Omega_{x,y})[(\alpha_{x,y}-1)+\displaystyle\sum_{t \in \mathcal{T}}  \displaystyle\sum_{d \in \mathcal{D} \setminus 1}                   \xi_o(O_{t,d-1,x},O_{t,d,y})]  - \lambda(\displaystyle\sum_{k \in \mathcal{S}}\Omega_{x,k}-1) + C.
\end{align*}
where C is a constant. 

Taking the derivative with respect to $\Omega_{x,y}$ and setting the equation equal to zero yields:
\begin{center}
$\dfrac{(\alpha_{x,y} -1 + \displaystyle\sum_{t \in \mathcal{T}}  \displaystyle\sum_{d \in \mathcal{D} \setminus 1}                   \xi_o(O_{t,d-1,x},O_{t,d,y})}{\Omega_{x,y}}  - \lambda = 0$\\
\end{center}
Rearranging,
\begin{center}

$\dfrac{(\alpha_{x,y} -1 + \displaystyle\sum_{t \in \mathcal{T}}  \displaystyle\sum_{d \in \mathcal{D} \setminus 1}                   \xi_o(O_{t,d-1,x},O_{t,d,y})}{\lambda}   = \Omega_{x,y}$\\

\end{center}
Taking the derivative of $F$ to be maximized with respect to $\lambda$ and setting the equation equal to zero yields:
\begin{center}
$(\displaystyle\sum_{k \in \mathcal{S}}\Omega_{x,k})-1 = 0$\\
\end{center}
Substituting the expression for $\Omega_{x,y}$  above into the equation immediately allows one to solve for $\lambda$, and thus for the update calculation for $\Omega_{x,y}$. The updates for each parameter can be found similarly and are

\begin{center}

$\Omega_{k,j}=\dfrac{   (\alpha_{k,j} -1) +             \displaystyle\sum_{t \in \mathcal{T}}   \displaystyle\sum_{d \in \mathcal{D} \setminus 1}    \xi_o(O_{t,d-1,k},O_{t,d,j})                                     }{                         \displaystyle\sum_{i \in \mathcal{S}}[   (\alpha_{k,i} -1) +   \displaystyle\sum_{t \in \mathcal{T}}   \displaystyle\sum_{d \in \mathcal{D} \setminus 1}    \xi_o(O_{t,d-1,k},O_{t,d,i})                   ]}$\\

$\pi_ k= \dfrac{\displaystyle\sum_{t \in \mathcal{T}}[\gamma_o(O_{t,1,k})]}                          {\displaystyle\sum_{i \in \mathcal{S}}[\displaystyle\sum_{t \in \mathcal{T}}[\gamma_o(O_{t,1,i})]]}$\\

$\rho_ k= \dfrac{\displaystyle\sum_{t \in \mathcal{T}}[\gamma_d(D_{t,1,k})]}                          {\displaystyle\sum_{i \in \mathcal{S}}[\displaystyle\sum_{t \in \mathcal{T}}[\gamma_d(D_{t,1,i})]]}$\\

$\Delta_{k,j}=\dfrac{      (\alpha_{k,j} - 1)   +       \displaystyle\sum_{t \in \mathcal{T}}   \displaystyle\sum_{d \in \mathcal{D} \setminus 1}     \xi_d(D_{t,d-1,k},D_{t,d,j})                                     }{                         \displaystyle\sum_{i \in \mathcal{S}}[   (\alpha_{k,i} - 1)  +    \displaystyle\sum_{t \in \mathcal{T}}   \displaystyle\sum_{d \in \mathcal{D} \setminus 1}     \xi_d(D_{t,d-1,k},D_{t,d,i})                   ]}$\\

$\Psi_{k,j,g}= \dfrac{ \beta_g + \displaystyle\sum_{h \in \mathcal{T}}   \displaystyle\sum_{a \in \mathcal{T}}    \displaystyle\sum_{d \in \mathcal{D}}                       (M_{h,a,d,g})\zeta(O_{h,d,k},D_{a,d,j}) }                                            { \displaystyle\sum_{g \in \mathcal{G}}[\beta_g + \displaystyle\sum_{h \in \mathcal{T}}   \displaystyle\sum_{a \in \mathcal{T}}    \displaystyle\sum_{d \in \mathcal{D}}                        (I_{h,a,d}) \zeta(O_{h,d,k},D_{a,d,j}) ]} $

$\Gamma_{k,j,g}= \dfrac{ \phi_g + \displaystyle\sum_{h \in \mathcal{T}}   \displaystyle\sum_{a \in \mathcal{T}}    \displaystyle\sum_{d \in \mathcal{D}}                      (N_{a,h,d,g})\zeta(O_{a,d,k},D_{h,d,j}) }                                            { \displaystyle\sum_{g \in \mathcal{G}}[\phi_g + \displaystyle\sum_{h \in \mathcal{T}}   \displaystyle\sum_{a \in \mathcal{T}}    \displaystyle\sum_{d \in \mathcal{D}}                      (I_{h,a,d}) \zeta(O_{a,d,k},D_{h,d,j}) ]} $

\end{center}

The MAP updates above were used to do update parameters iteratively.\footnote{In our implementation, we applied an additional condition to the goal factors $\Psi$ and $\Gamma$ that leads them to no longer have an analytic form. We enforced the condition that higher state numbers corresponded to higher strength. We did this by replacing the analytic update formula for $\Psi$ above with an interior-point optimization in which we found all optimal parameters $\Psi_{k,j,g}$ that maximized the objective likelihood function subject to several conditions. First, each row of the CPT was required to sum to one. Second, the expected number of goals scored by an offense at a higher strength was required to be higher than the expected number of goals scored by an offense at a lower strength, and that the expected number of goals allowed by a defense at a higher strength was required to be lower than the expected number of goals scored by a defense at a lower strength. A similar process was followed to calculate $\Gamma$.} To initialize the model, a random setting of the parameters was chosen. Thereafter the following steps were followed. First, belief propagation was used to compute the conditional distribution of the $\mathcal{Z}$ given the observed $\mathcal{X}$, selected hyperparameters $\Lambda$, and current setting of the $\theta$. Because of the loops in the factor graph, loopy belief propagation was used, which required selecting a message passing schedule and repeatedly distributing messages through the network.\footnote{With the selected schedule, 20 cycles of message passing were found to be sufficient.} Second, messages from belief propagation were used to calculate the marginal distribution over each individual element of $\mathcal{Z}$ and over relevant pairs of variables in $\mathcal{Z}$, which yielded $\gamma$, $\xi$, and $\zeta$. Third, updated parameters $\pi$, $\rho$, $\Omega$, $\Delta$, $\Psi$, and $\Gamma$ were calculated according to the formulae above. The network was then updated with these new parameter values and a new iteration was begun. This process was repeated until convergence. 

\section{Experiments}

\begin{figure*}[ht]
\begin{center}\includegraphics[scale=0.40]{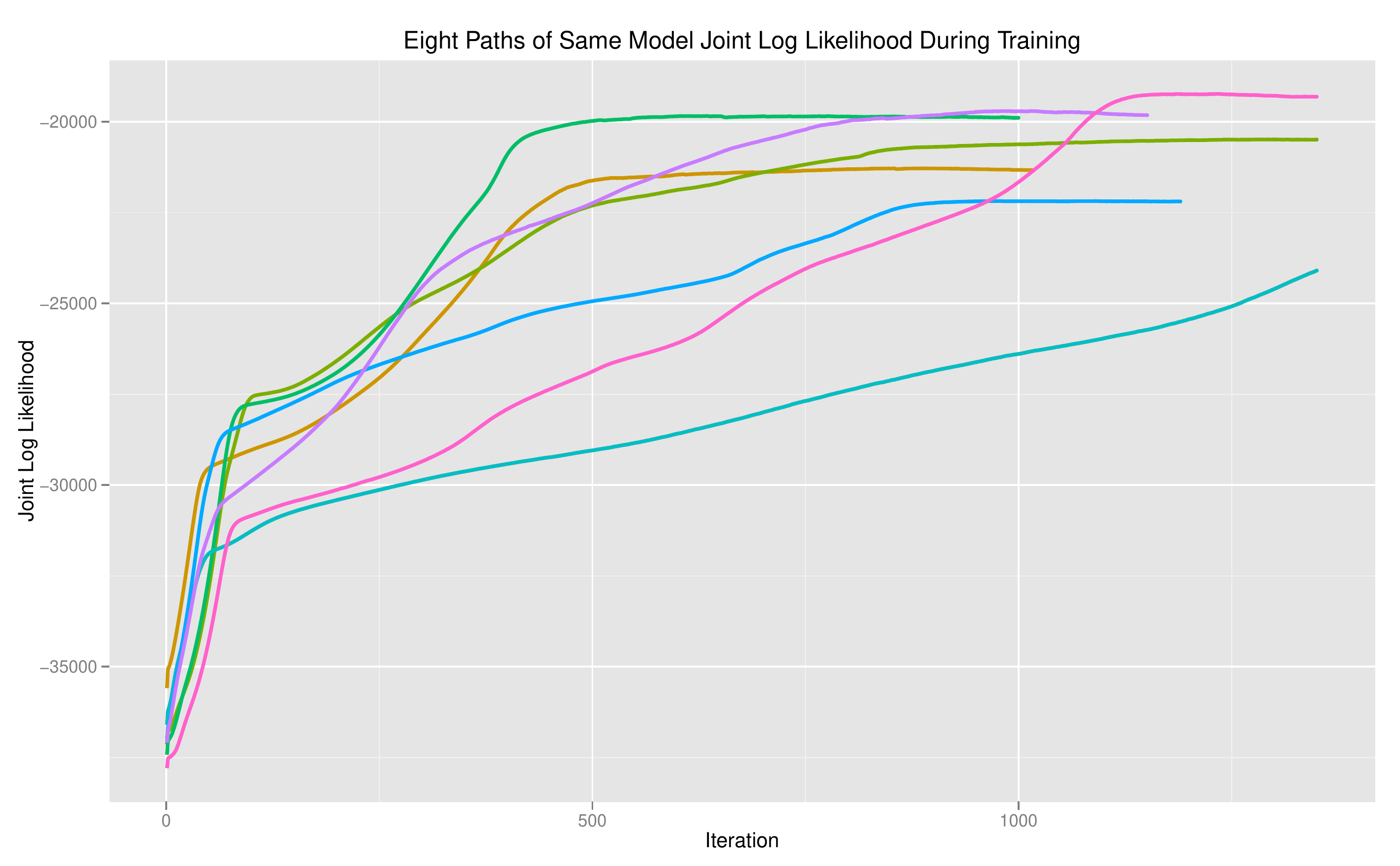}
\caption{Eight paths of joint log likelihood during training after starting at different initializations\label{fig:eight_paths}}
\end{center}
\end{figure*}

English Premier League match data and corresponding William Hill sportsbook lines were obtained from football-data.co.uk \citep{FOOTBALLDATA}. The algorithm was implemented in C++. We were able to experiment with many different combinations of strength states and regularization parameters to evaluate which ones were optimal. The non-convexity of the objective function proved to be a serious challenge in training the model. Given in Figure~\ref{fig:eight_paths} are eight paths of the joint log likelihood for the same case: training data, regularization parameters, and number of strength states employed are identical in each instance. The only difference between these cases is the randomization of the initial parameter settings. 
As is evident above, each of the eight initializations resulted in different parameter estimates. The non-convexity of the objective function leads to inconsistent parameter estimation and requires repeated runs and cross-validation to ensure that the computed parameters are reasonable. This is an issue commonly encountered with EM-based approaches. That said, the learned models were subjectively similar with respect to learned offensive and defensive strength estimates. We show results from the models with the best cross-validated predictive performance.

\begin{figure*}[ht]
\begin{center}
\includegraphics[scale=0.21]{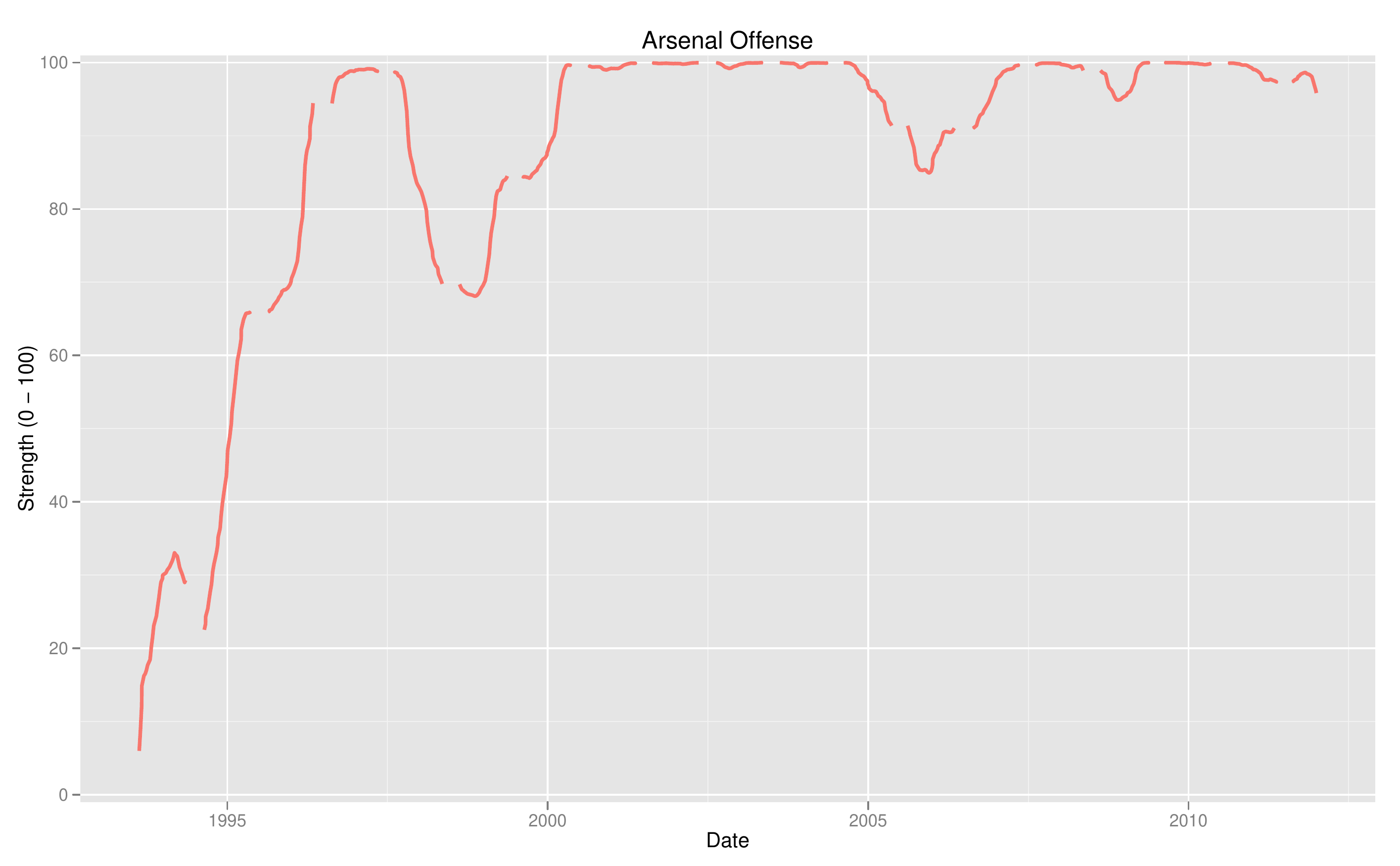}
\includegraphics[scale=0.21]{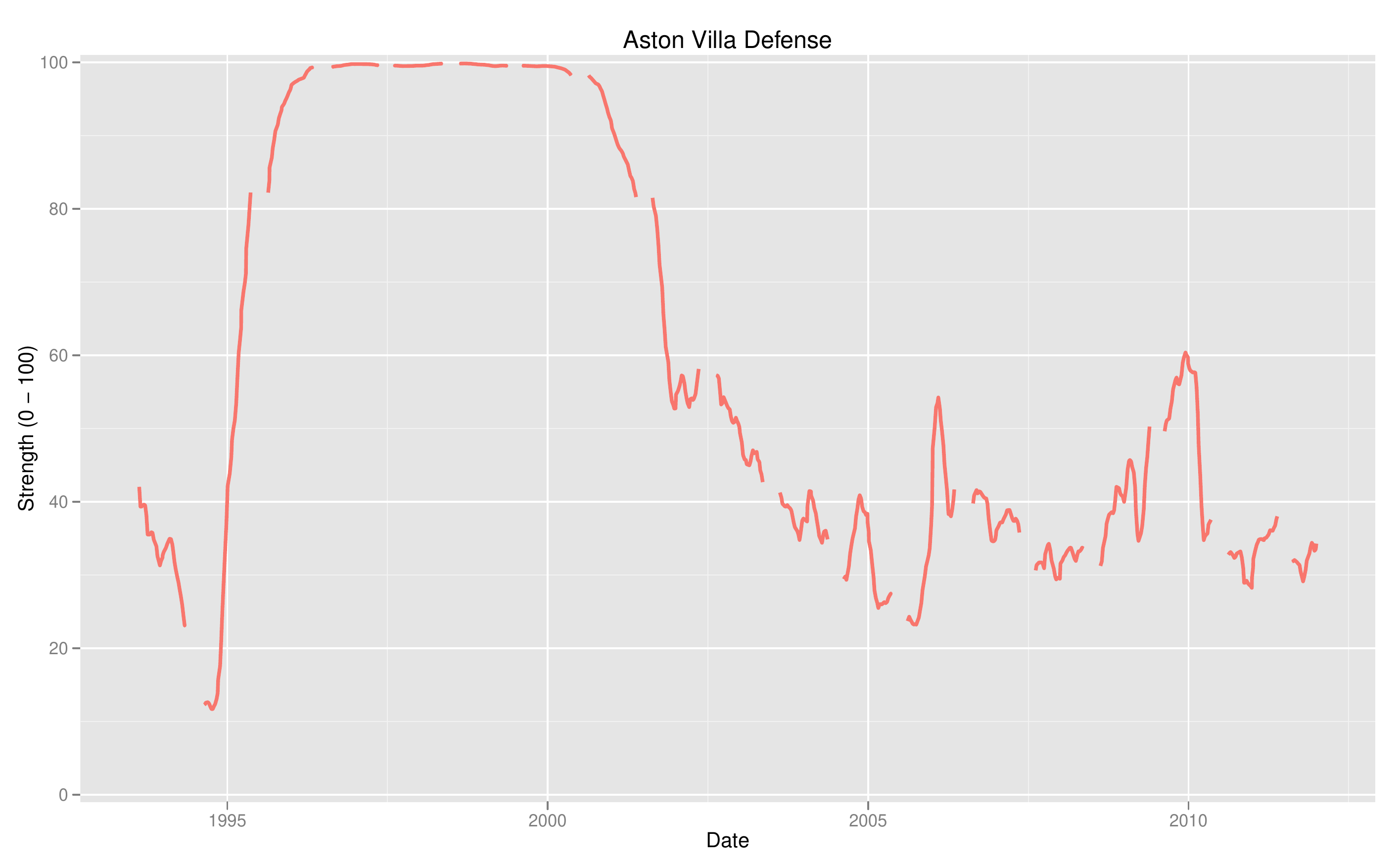}
\includegraphics[scale=0.21]{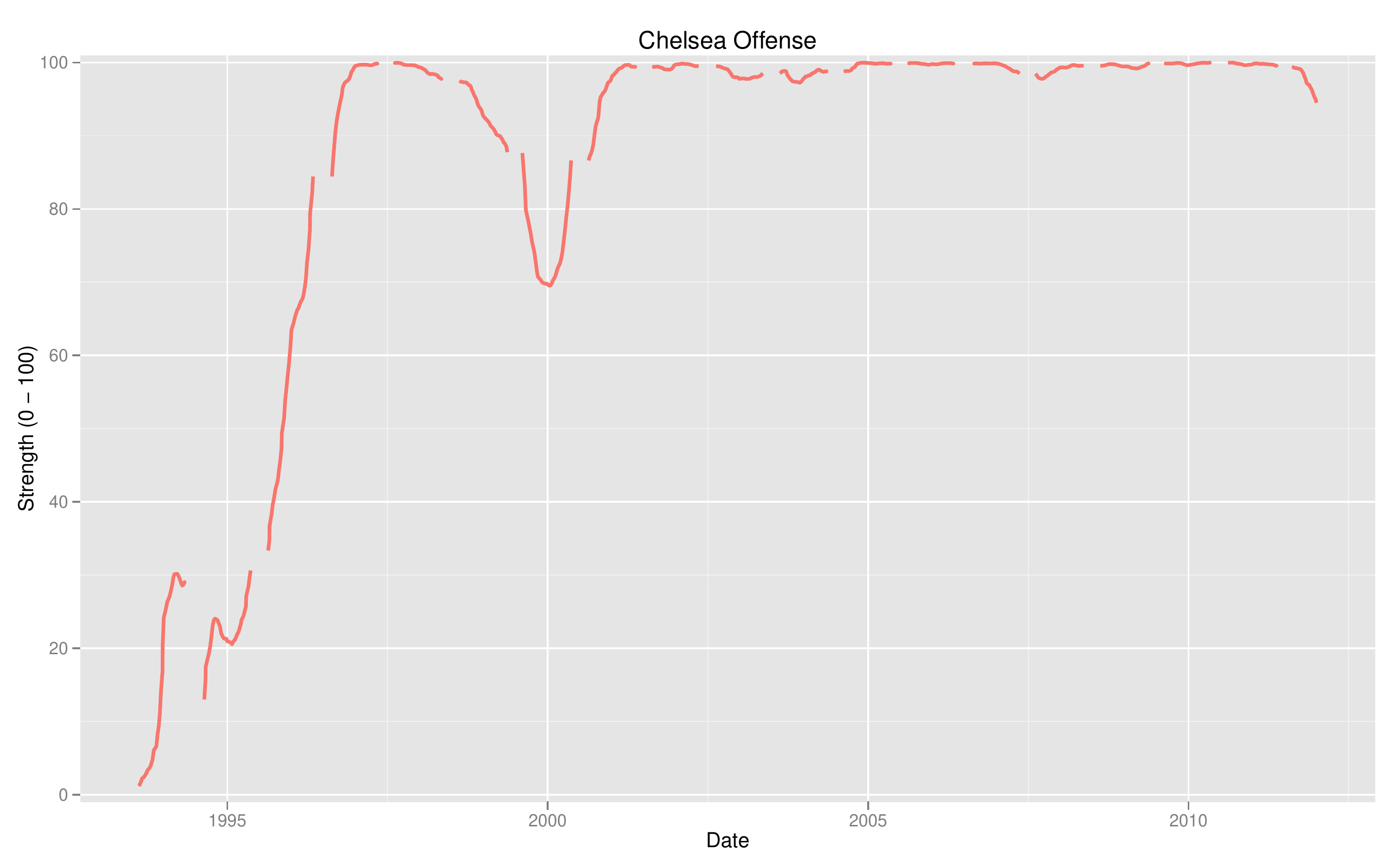}
\includegraphics[scale=0.21]{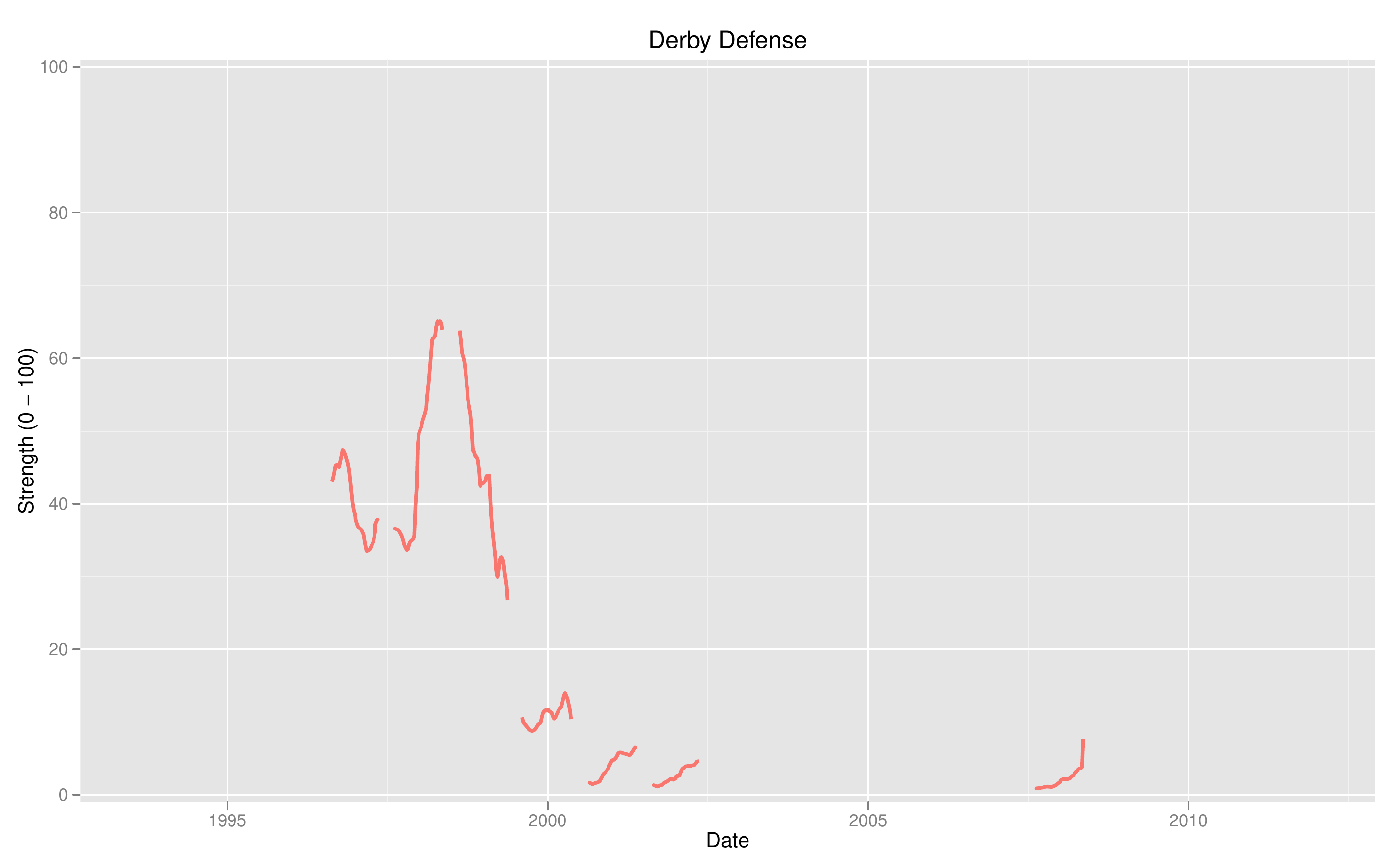}
\includegraphics[scale=0.21]{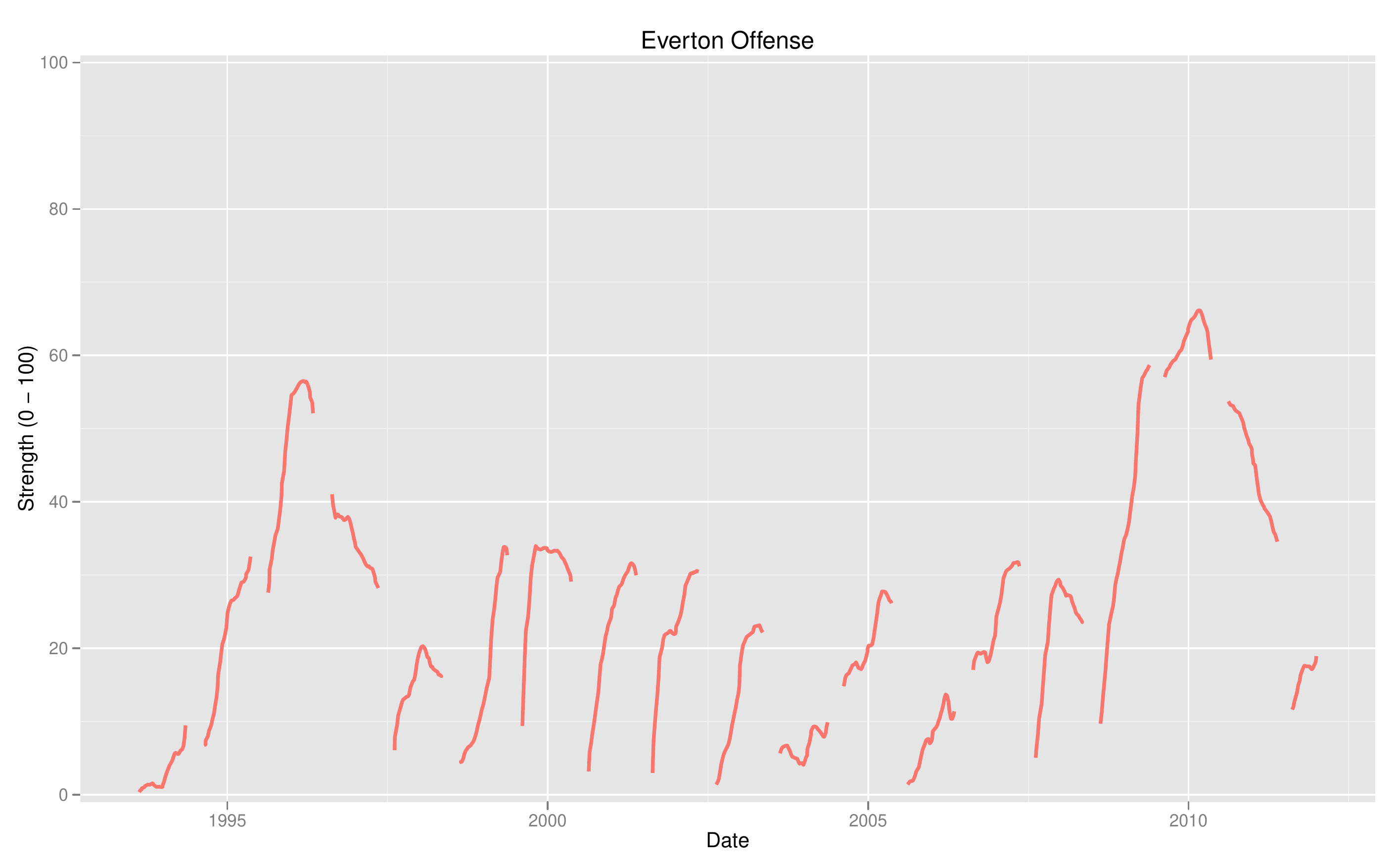}
\includegraphics[scale=0.21]{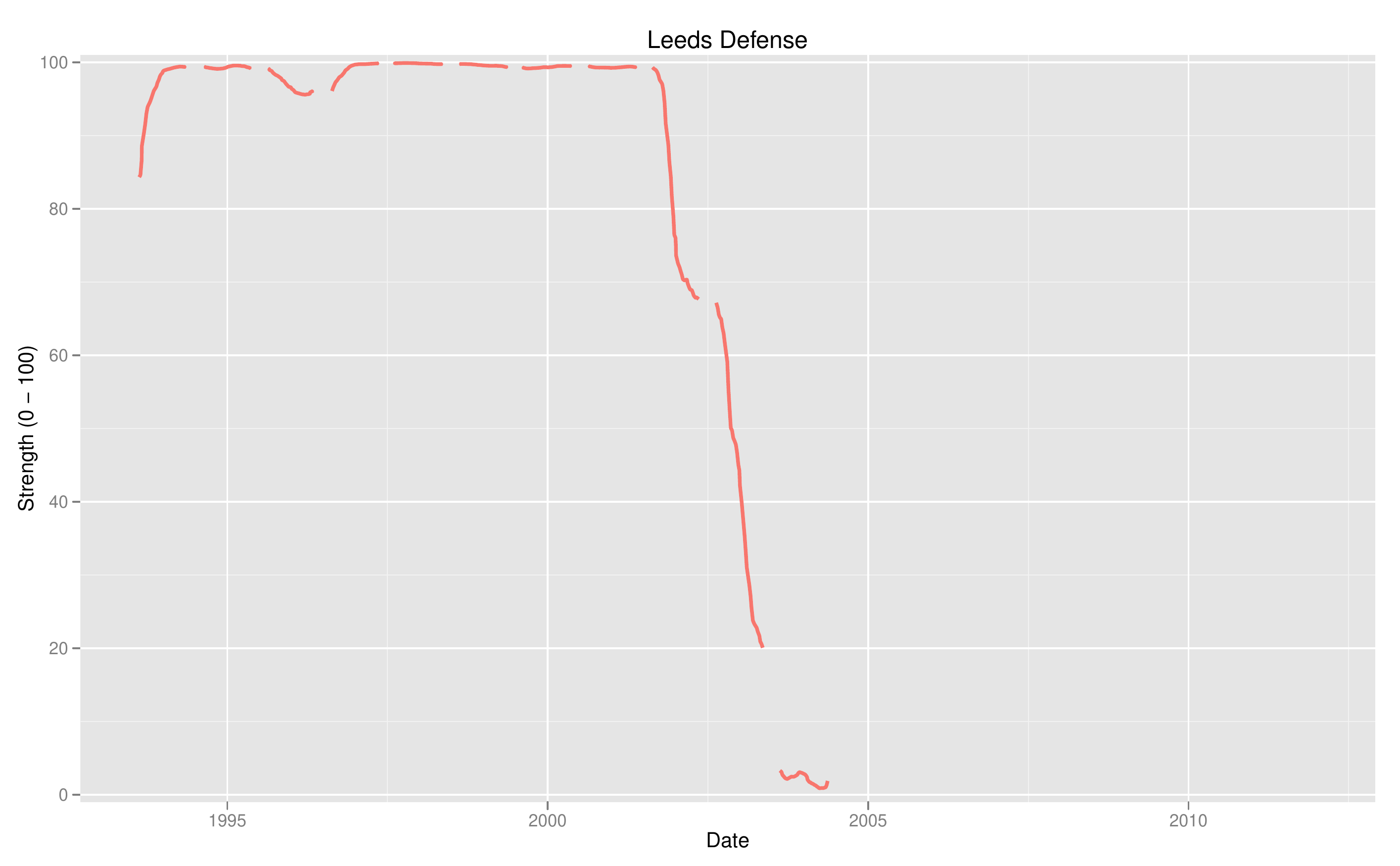}
\includegraphics[scale=0.21]{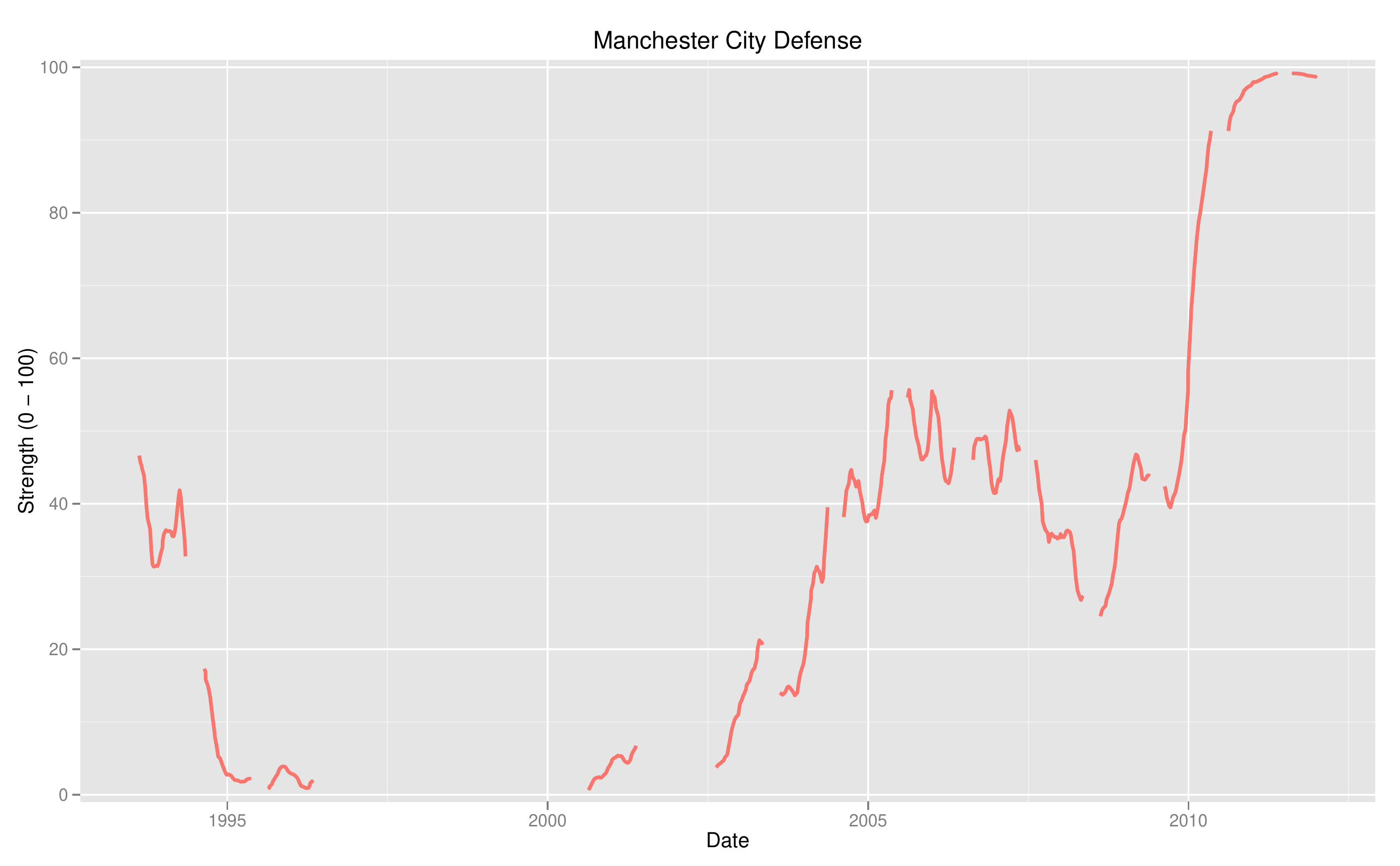}
\includegraphics[scale=0.21]{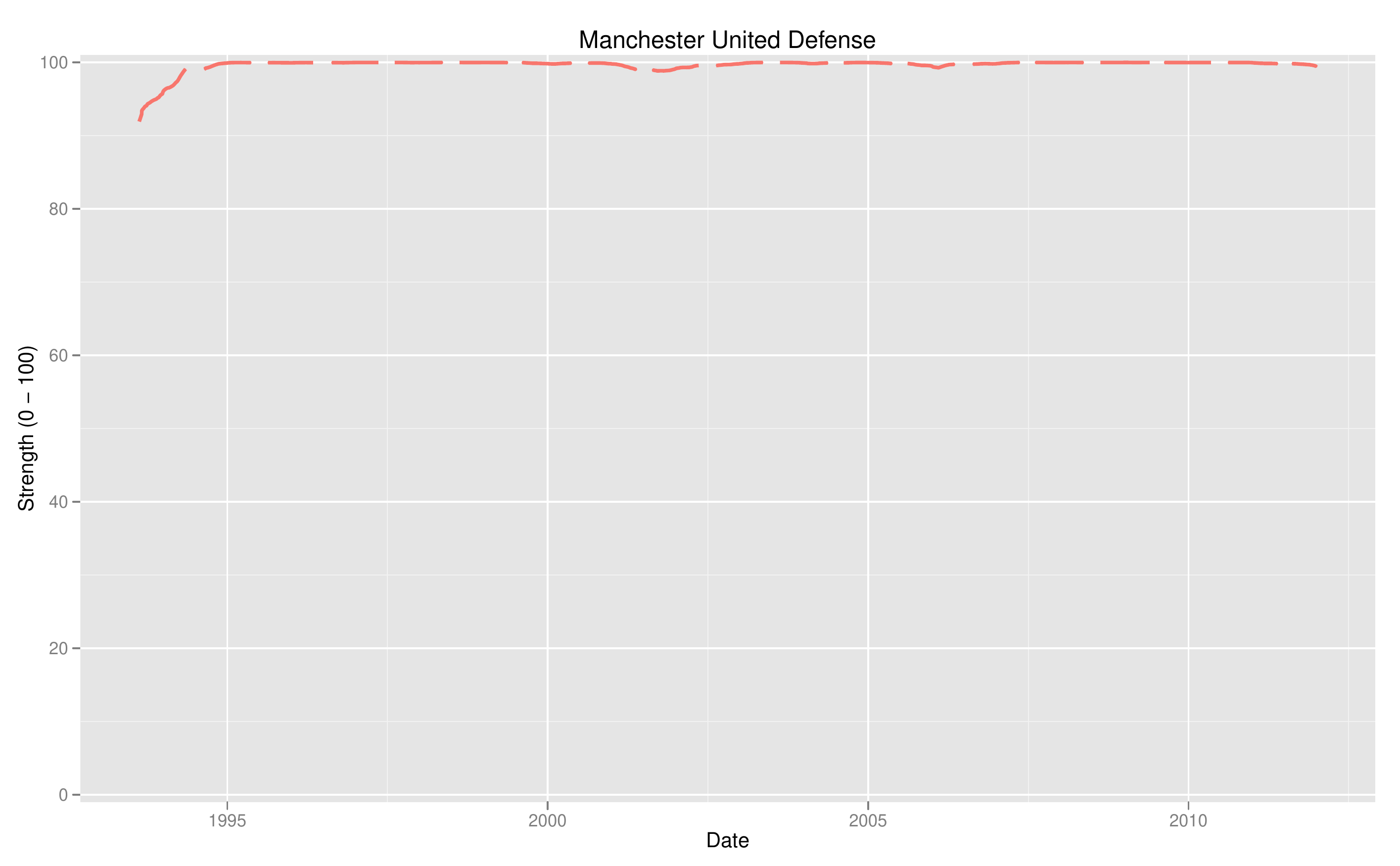}
\includegraphics[scale=0.21]{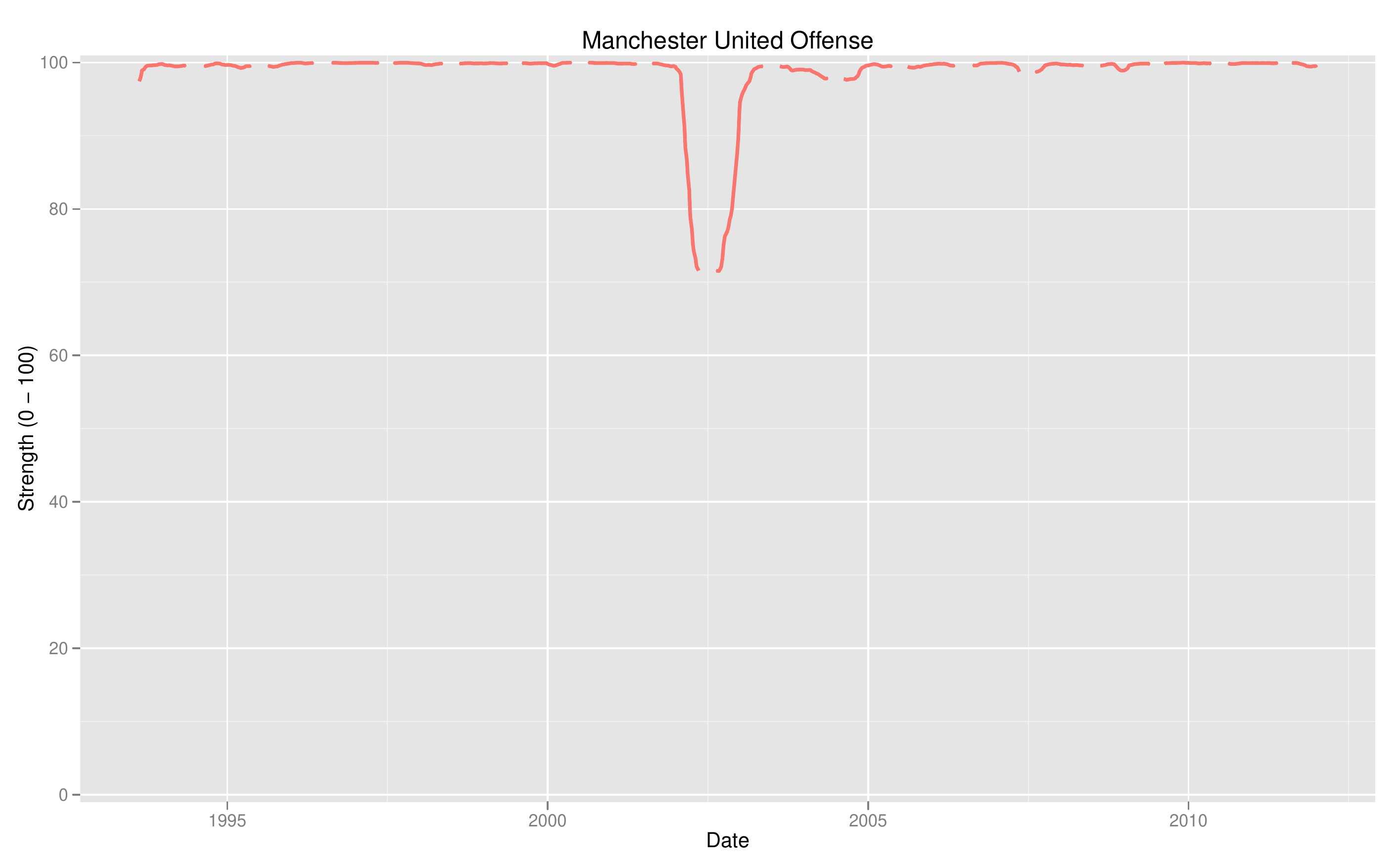}
\includegraphics[scale=0.21]{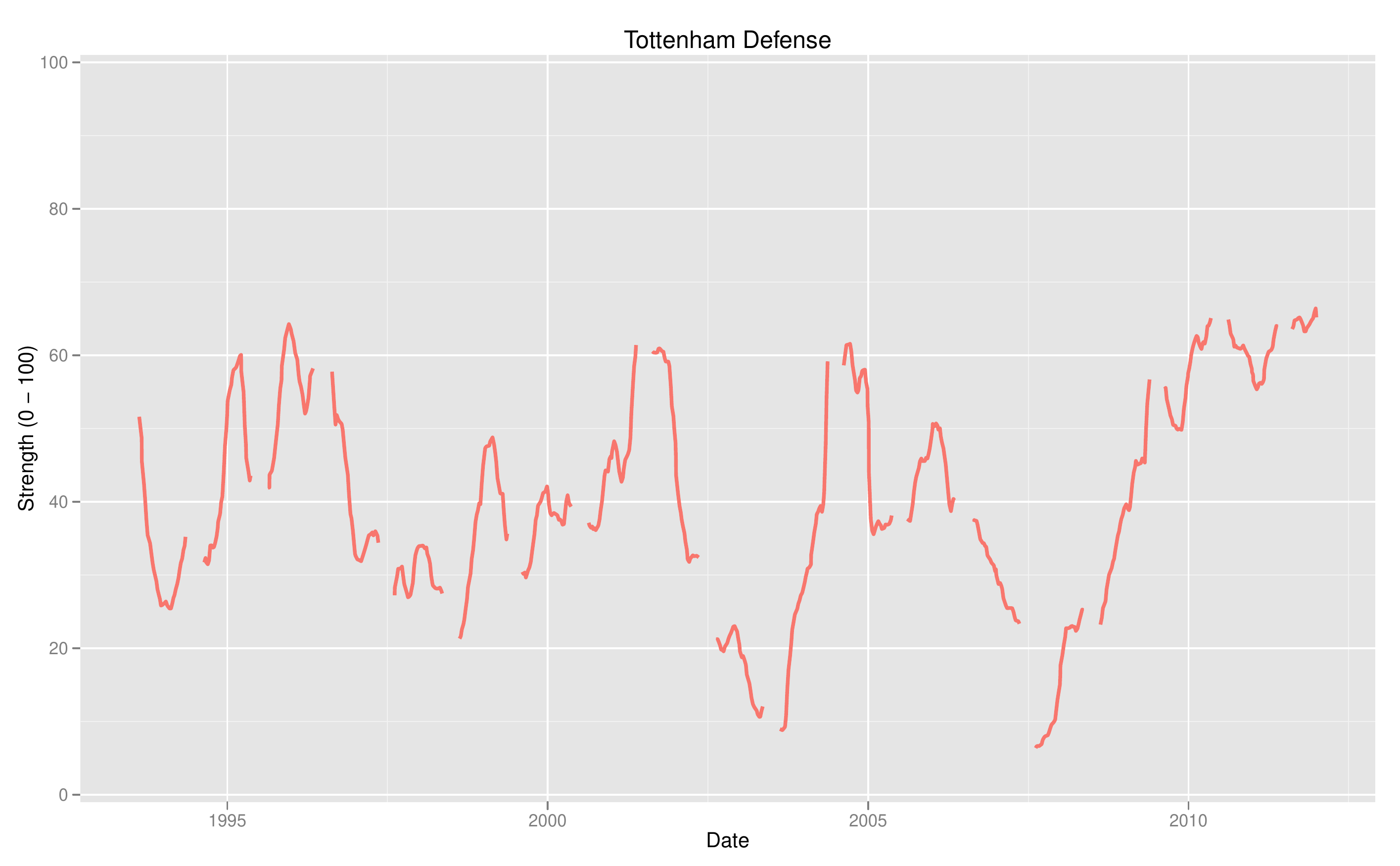}
\caption{Examples of inferred offensive and defensive strengths over time. For the purposes of plotting, an expectation was taken over four discrete states (0-3), and this expectation was rescaled to a strength ranging from 0 to 100. Breaks in lines represent time between seasons. \label{fig:example_graphs}}
\end{center}
\end{figure*}

A number of graphs of offensive and defensive strength over time as inferred by the model are given in Figure~\ref{fig:example_graphs}.
 
A number of the shifts in the above graphs can be corroborated with external events that significantly affected the team. As an example, Leeds faced serious financial issues after the 2001 season. These  difficulties forced them to trade away a number of their highly prized defenders, including Rio Ferdinand and Jonathan Woodgate \citep{LEEDSTRADE}. These events are reflected in the rapid decline of their defense in this period. 

The final parameters used to generate the inferred team strengths above are given in Figures~\ref{fig:parameters1} and~\ref{fig:parameters2}. We used 4 strength states, and as regularization parameters chose 236 goal pseudocounts for $\Psi$ and $\Gamma$ and 87 pseudocounts for $\Omega$ and $\Delta$.

\begin{figure*}[ht]
\begin{center}
\begin{filecontents*}{pi.csv}
Parameter,Estimate
$\pi_{0}$,0.817
$\pi_{1}$,0.034
$\pi_{2}$,0.082
$\pi_{3}$,0.066
\end{filecontents*}
\csvautotabular{pi.csv}
\begin{filecontents*}{rho.csv}
Parameter,Estimate
$\rho_{0}$,0.476
$\rho_{1}$,0.07
$\rho_{2}$,0.372
$\rho_{3}$,0.082
\end{filecontents*}
\csvautotabular{rho.csv}\\

$\,$\\

Indexed values of $\Omega_{i,j}$:\\
\begin{filecontents*}{omega.csv}
,j=0,1,2,3
i=0,0.957,0.043,0.000,0.000
1,0.021,0.970,0.009,0.000
2,0.000,0.014,0.974,0.011
3,0.000,0.000,0.008,0.992
\end{filecontents*}
\csvautotabular{omega.csv}

Indexed values of $\Delta_{i,j}$: \\
\begin{filecontents*}{delta.csv}
,j=0,1,2,3
i=0,0.979,0.021,0,0
1,0.009,0.946,0.045,0
2,0,0.09,0.906,0.005
3,0,0,0.002,0.998
\end{filecontents*}
\csvautotabular{delta.csv}

\caption{Parameters estimated for teams in the English Premier League, 8/14/1993 - 12/31/2011.  $\pi$ and $\rho$ encode the initial probability of a team beginning the period in different strength states. Few teams are excellent, and as such little mass is put on the top strength state (3) while more is put on the lowest strength state (0). $\Omega$ and $\Delta$ encode transitions, and we can see that teams usually stay in the same state week-to-week.\label{fig:parameters1}}
\end{center}
\end{figure*}

\begin{figure*}[ht]
\begin{center}
$\,$\\

\begin{filecontents*}{psi_and_gamma.csv}
a:,0,1,2,3,,h:,0,1,2,3
$\Psi_{0,a,0}$,0.245,0.355,0.225,0.396,,$\Gamma_{0,h,0}$,0.244,0.397,0.501,0.570
$\Psi_{0,a,1}$,0.353,0.235,0.498,0.400,,$\Gamma_{0,h,1}$,0.442,0.384,0.288,0.230
$\Psi_{0,a,2}$,0.224,0.255,0.214,0.163,,$\Gamma_{0,h,2}$,0.216,0.176,0.155,0.154
$\Psi_{0,a,3}$,0.102,0.125,0.036,0.039,,$\Gamma_{0,h,3}$,0.088,0.030,0.039,0.038
$\Psi_{0,a,4}$,0.076,0.030,0.027,0.003,,$\Gamma_{0,h,4}$,0.011,0.014,0.017,0.008
$\Psi_{1,a,0}$,0.125,0.113,0.380,0.389,,$\Gamma_{1,h,0}$,0.339,0.376,0.368,0.381
$\Psi_{1,a,1}$,0.295,0.407,0.204,0.329,,$\Gamma_{1,h,1}$,0.268,0.346,0.389,0.533
$\Psi_{1,a,2}$,0.334,0.307,0.227,0.203,,$\Gamma_{1,h,2}$,0.205,0.182,0.190,0.026
$\Psi_{1,a,3}$,0.132,0.139,0.116,0.061,,$\Gamma_{1,h,3}$,0.139,0.054,0.038,0.052
$\Psi_{1,a,4}$,0.114,0.033,0.073,0.019,,$\Gamma_{1,h,4}$,0.049,0.042,0.015,0.009
$\Psi_{2,a,0}$,0.123,0.168,0.289,0.297,,$\Gamma_{2,h,0}$,0.149,0.307,0.351,0.420
$\Psi_{2,a,1}$,0.293,0.440,0.331,0.301,,$\Gamma_{2,h,1}$,0.447,0.359,0.319,0.359
$\Psi_{2,a,2}$,0.220,0.124,0.172,0.246,,$\Gamma_{2,h,2}$,0.284,0.137,0.191,0.156
$\Psi_{2,a,3}$,0.282,0.093,0.110,0.117,,$\Gamma_{2,h,3}$,0.106,0.130,0.062,0.060
$\Psi_{2,a,4}$,0.083,0.176,0.098,0.039,,$\Gamma_{2,h,4}$,0.014,0.067,0.078,0.005
$\Psi_{3,a,0}$,0.069,0.069,0.180,0.177,,$\Gamma_{3,h,0}$,0.211,0.190,0.242,0.327
$\Psi_{3,a,1}$,0.195,0.203,0.256,0.401,,$\Gamma_{3,h,1}$,0.229,0.323,0.387,0.364
$\Psi_{3,a,2}$,0.288,0.306,0.333,0.258,,$\Gamma_{3,h,2}$,0.212,0.300,0.232,0.206
$\Psi_{3,a,3}$,0.172,0.212,0.193,0.116,,$\Gamma_{3,h,3}$,0.260,0.101,0.107,0.075
$\Psi_{3,a,4}$,0.275,0.211,0.038,0.049,,$\Gamma_{3,h,4}$,0.088,0.086,0.031,0.027

\end{filecontents*}
\csvautotabular{psi_and_gamma.csv}

\caption{Parameters estimated for teams in the English Premier League, 8/14/1993 - 12/31/2011. $\Psi$ and $\Gamma$ encode goal factors, in which one can see that stronger offensive teams tend to score more goals while stronger defensive teams allow fewer goals.\label{fig:parameters2}}
\end{center}
\end{figure*}

Our model aims to provide an interpretable measurement of a team's strength over time. We do this by building a generative model for goals scored that relies upon the latent strengths of the two competing teams. After parameter estimation and inference of a distribution over historical team strength is completed, we are left with a full generative model for the English Premier League. 

We could additionally use this model to predict the outcome of future contests, and we experimented with this potential use of the model. We evaluated the model's out-of-sample, real-time predictive performance against two competing methods. The first is the implied win, draw, and loss percentage inferred from sports betting lines posted by William Hill, a popular London sportsbook. The second is the prediction made by an Elo model with parameters optimized on the same training data we use to fit our model. We applied our model to data from the English Premier League.

\begin{figure*}[ht]
\begin{center}
\includegraphics[scale=0.40]{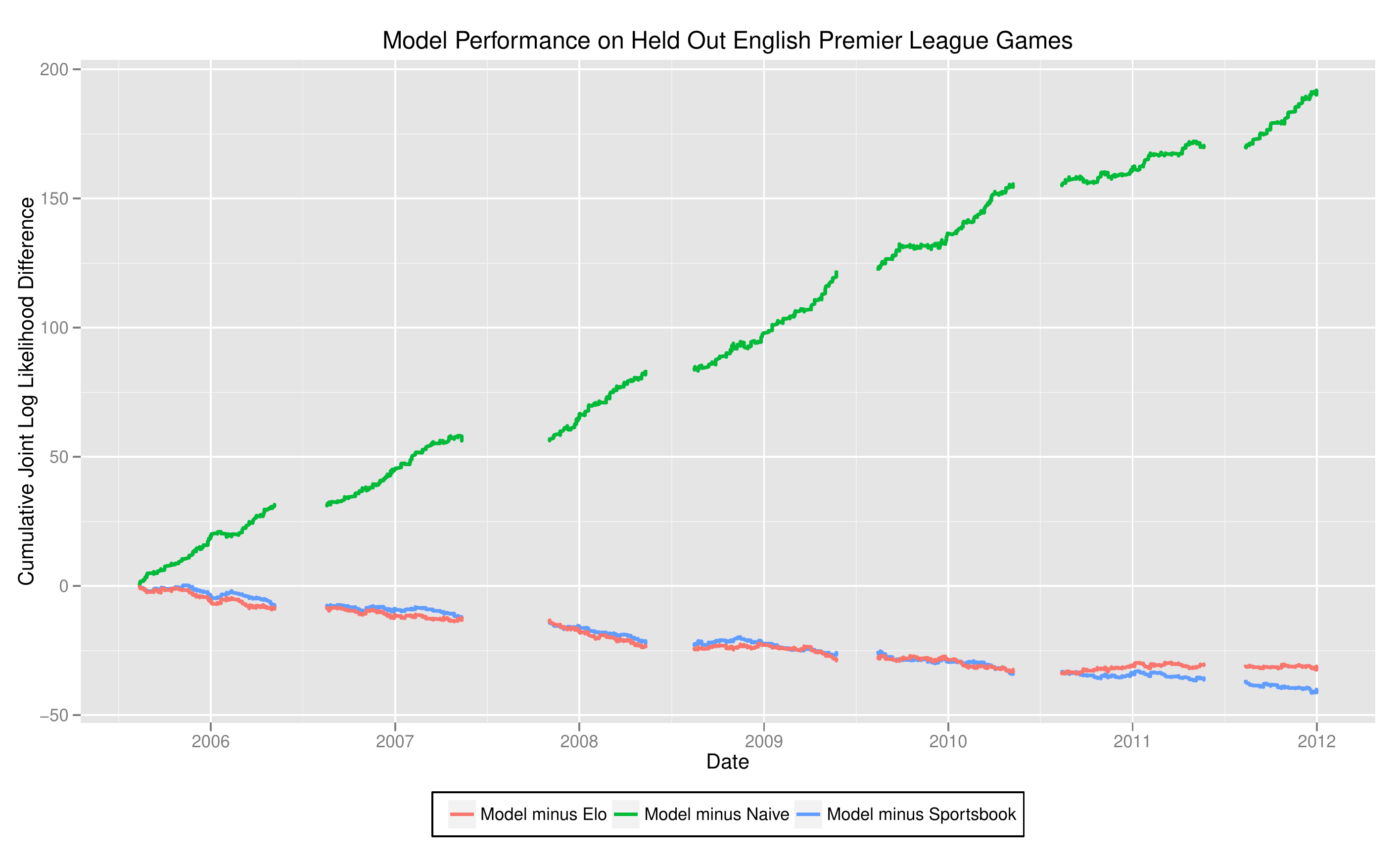}
\caption{Model performance on held out English Premier League Games, 2005-2012. The model was initially trained using data through the end of the 2004-2005 season. For every subsequent week, scoring predictions were made and these were translated into a probability of win, draw, and loss. These were compared with implied win, draw, loss probabilities from William Hill, Elo, and a naive method that assumed an optimal constant home win, draw, and loss percentage. The cumulative net log likelihood the model placed on the actual outcome of the game compared to that of each of these three methods is depicted above. After each week's predictions had been made, the model was allowed to incorporate that week's data into future predictions by completing ten iterations of the coordinate ascent procedure before proceeding onto the next week. \label{fig:eplholdout}}
\end{center}
\end{figure*}

Figure~\ref{fig:eplholdout} gives our predictive results.

On a large sample of English Premier League data,  the model's predictions added substantial information versus a naive approach in which every game is estimated to have identical home win, draw, and loss probabilities. The model underformed both Elo and expert human linemakers, however.

\section{Discussion}

We believe that estimations of historical team strength could have use in a variety of different contexts. Sports enthusiasts might like to examine inferred trends in the strength of their favorite teams when visiting sports news websites. Recreational sports bettors might enjoy comparing team strengths before placing their bets. Coaches and strategists might find these inferred strengths useful decision-support tools: they could compare how a team's inferred strength evolved after known changes to the team were made. In general, these inferred strengths offer a way of seeing something that cannot be directly seen: how skilled a team's offense or defense truly is. Inferred strengths can be used to remove the guesswork from making an assessment about whether or not a team's skill is likely to have truly changed after a particularly good or bad run of performance. 

This model could be applied to a number of other sports with low, discrete scores, such as ice hockey, field hockey, and lacrosse. The strengths inferred by this model could also be used as an additional feature in models designed to predict the outcome of matches. As this is an original approach to modeling soccer, we believe these inferred strengths may contain new information that could add to an existing predictive model. We believe that real-time predictions based exclusively on this model are naturally limited by the discrete state-space modeling of latent strength, but one who wished to make the most accurate possible predictions of games could extend this modeling framework into a continuous space through the use of Gaussian strength states and Poisson-parameterized goal emission factors.

\begin{figure*}[ht]
\begin{center}
\includegraphics[scale=0.35]{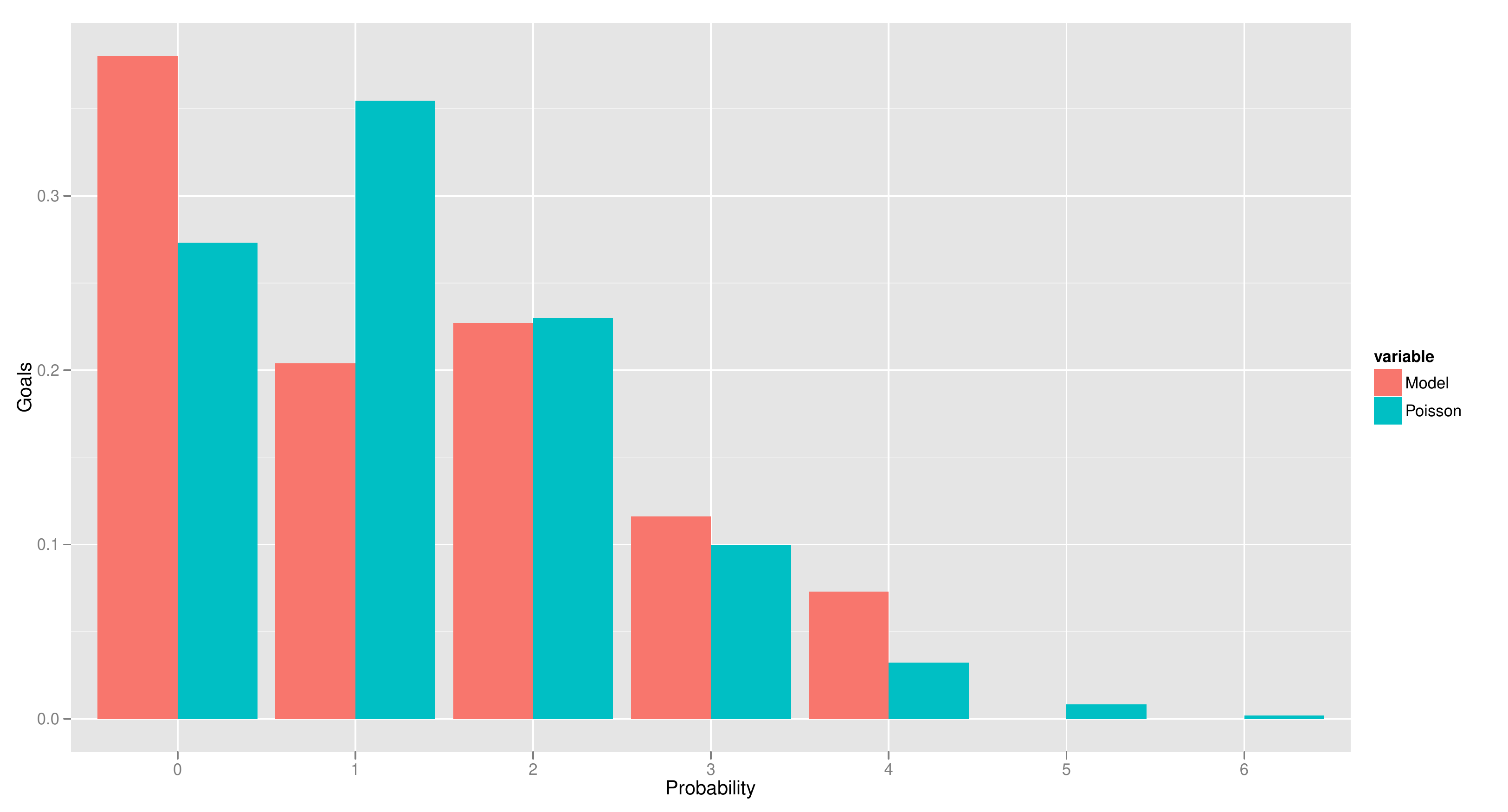}
\caption{Comparison of conditional probabilities encoded in $\Psi_{1,2,g}$ and matched Poisson distribution.\label{fig:poissonfit}}
\end{center}
\end{figure*}

The parameters estimated by this model may themselves be of interest to sports analysts. Our goal emission factors were very flexibly parameterized and were allowed to discover the discrete conditional distributions over goals that best fit the observed data. The Poisson distribution is commonly used to model soccer, but we did not impose that restriction on our goal emission factors. In fact, a number of the conditional distributions we discovered deviated noticeably from a Poisson distribution. For example, we can examine the distribution over home goals given by $\Psi_{1,2,g}$, when a relatively weaker home offense challenges a relatively stronger away defense. The distribution over goals as estimated by the model and by a comparable Poisson distribution (with $\lambda$ selected to match the expectation over this discrete distribution) is given in Figure \ref{fig:poissonfit}. There is a substantial difference in the shape of these two distributions, particularly in the amount of mass each assigns to zero and one goals. 

\begin{small}
\bibliographystyle{plainnat}
\bibliography{sample} 

\begin{thebibliography}{11}
\providecommand{\natexlab}[1]{#1}
\providecommand{\url}[1]{\texttt{#1}}
\expandafter\ifx\csname urlstyle\endcsname\relax
  \providecommand{\doi}[1]{doi: #1}\else
  \providecommand{\doi}{doi: \begingroup \urlstyle{rm}\Url}\fi

\bibitem[Bishop(2006)]{Bishop:2006:PRM:1162264}
Christopher~M. Bishop.
\newblock \emph{Pattern Recognition and Machine Learning (Information Science
  and Statistics)}.
\newblock Springer-Verlag New York, Inc., Secaucus, NJ, USA, 2006.
\newblock ISBN 0387310738.

\bibitem[Buchdahl(2013)]{FOOTBALLDATA}
Joseph Buchdahl.
\newblock {D}ata {F}iles: {E}ngland {@ONLINE}, March 2013.
\newblock URL \url{http://www.football-data.co.uk/englandm.php}.

\bibitem[Curiel(2012)]{ELOSOCCER}
Raoul S. da~Silva Curiel.
\newblock {W}orld {E}lo {F}ootball {E}lo {R}atings {@ONLINE}, August 2012.
\newblock URL \url{http://www.eloratings.net/system.html}.

\bibitem[Dempster et~al.(1977)Dempster, Laird, and
  Rubin]{Dempster77maximumlikelihood}
A.~P. Dempster, N.~M. Laird, and D.~B. Rubin.
\newblock Maximum likelihood from incomplete data via the {EM} algorithm.
\newblock \emph{JOURNAL OF THE ROYAL STATISTICAL SOCIETY, SERIES B},
  39\penalty0 (1):\penalty0 1--38, 1977.

\bibitem[ESPN(2003)]{LEEDSTRADE}
ESPN.
\newblock Leeds close to second major sale {@ONLINE}, January 2003.
\newblock URL \url{http://soccernet.espn.go.com/news/story?id=256367\&cc=5901}.

\bibitem[Frey and MacKay(1997)]{DBLP:conf/nips/FreyM97}
Brendan~J. Frey and David J.~C. MacKay.
\newblock {A} {R}evolution: {B}elief {P}ropagation in {G}raphs with {C}ycles.
\newblock In Michael~I. Jordan, Michael~J. Kearns, and Sara~A. Solla, editors,
  \emph{NIPS}. The MIT Press, 1997.
\newblock ISBN 0-262-10076-2.

\bibitem[Herbrich et~al.(2006)Herbrich, Minka, and
  Graepel]{DBLP:conf/nips/HerbrichMG06}
Ralf Herbrich, Tom Minka, and Thore Graepel.
\newblock True{S}kill$^{\mbox{\texttrademark}}$: {A} {B}ayesian {S}kill
  {R}ating {S}ystem.
\newblock In Bernhard Sch{\"o}lkopf, John~C. Platt, and Thomas Hoffman,
  editors, \emph{NIPS}, pages 569--576. MIT Press, 2006.
\newblock ISBN 0-262-19568-2.

\bibitem[{{North American SCRABBLE Players Association}}(2012)]{SCRAB}
{{North American SCRABBLE Players Association}}.
\newblock {R}ating {S}ystem {O}verview {@ONLINE}, August 2012.
\newblock URL \url{http://www.scrabbleplayers.org/w/Rating\textunderscore
  system\textunderscore overview}.

\bibitem[Pearl(1982)]{DBLP:conf/aaai/Pearl82}
Judea Pearl.
\newblock Reverend {B}ayes on {I}nference {E}ngines: {A} {D}istributed
  {H}ierarchical {A}pproach.
\newblock In David~L. Waltz, editor, \emph{AAAI}, pages 133--136. AAAI Press,
  1982.
\newblock ISBN 0-262-51051-2.

\bibitem[Silver(2009)]{Silver:2009:Online}
Nate Silver.
\newblock {A} {G}uide to {ESPN}'s {SPI} {R}atings {@ONLINE}, December 2009.
\newblock URL
  \url{http://soccernet.espn.go.com/world-cup/story/\textunderscore/id/4447078/ce/us/
  guide-espn-spi-ratings}.

\bibitem[Yedidia et~al.(2000)Yedidia, Freeman, and
  Weiss]{DBLP:conf/nips/YedidiaFW00}
Jonathan~S. Yedidia, William~T. Freeman, and Yair Weiss.
\newblock {G}eneralized {B}elief {P}ropagation.
\newblock In Todd~K. Leen, Thomas~G. Dietterich, and Volker Tresp, editors,
  \emph{NIPS}, pages 689--695. MIT Press, 2000.

\end{thebibliography}
\end{small}
\end{document}